\documentclass[journal]{IEEEtran}
\IEEEoverridecommandlockouts

\usepackage{amsmath,amsfonts}
\usepackage{array}
\usepackage{caption2}
\usepackage{textcomp}
\usepackage{stfloats}
\usepackage{url}
\usepackage{verbatim}
\usepackage{graphicx}
\usepackage{cite}

\usepackage{amssymb}
\usepackage{enumitem}
\usepackage{booktabs}
\usepackage{subfigure}
\usepackage{rotating, multirow}
\usepackage[ruled]{algorithm2e}
\usepackage{bbm}
\usepackage{xcolor}
\usepackage{balance}

\usepackage[capitalize]{cleveref}

\newcommand{\tabincell}[2]{\begin{tabular}{@{}#1@{}}#2\end{tabular}}

\AtBeginDocument{%
  \providecommand\BibTeX{{%
    \normalfont B\kern-0.5em{\scshape i\kern-0.25em b}\kern-0.8em\TeX}}}

\ifCLASSINFOpdf

\else

\fi

\begin{document}

\title{Effective and Robust Adversarial Training against \\ Data and Label Corruptions}

\author{Peng-Fei Zhang, Zi Huang, Xin-Shun Xu, Guangdong Bai
\thanks{This work is partially supported by Australian Research Council Discovery Project (DP230101196, CE200100025).}
\thanks{Peng-Fei Zhang, Zi Huang and Guangdong Bai are with the School of Information Technology \& Electrical Engineering, the University of Queensland, email: mima.zpf@gmail.com, huang@itee.uq.edu.au and g.bai@uq.edu.au.}
\thanks{Xin-Shun Xu is with the School of Software, Shandong University, Jinan, China, email: xuxinshun@sdu.edu.cn.}
}

\markboth{Journal of \LaTeX\ Class Files,~Vol.~14, No.~8, August~2015}%
{Shell \MakeLowercase{\textit{et al.}}: Bare Demo of IEEEtran.cls for IEEE Journals}

\maketitle

\begin{abstract}
   
   Corruptions due to data perturbations and label noise are prevalent in the datasets from unreliable sources, which poses significant threats to model training. Despite existing efforts in developing robust models, current learning methods commonly overlook the possible co-existence of both corruptions, limiting the effectiveness and practicability of the model.

    In this paper, we develop an Effective and Robust Adversarial Training (ERAT) framework to simultaneously handle two types of corruption (i.e., data and label) without prior knowledge of their specifics. We propose a hybrid adversarial training surrounding multiple potential adversarial perturbations, alongside a semi-supervised learning based on class-rebalancing sample selection to enhance the resilience of the model for dual corruption. On the one hand, in the proposed adversarial training, the perturbation generation module learns multiple surrogate malicious data perturbations by taking a DNN model as the victim, while the model is trained to maintain semantic consistency between the original data and the hybrid perturbed data. It is expected to enable the model to cope with unpredictable perturbations in real-world data corruption. On the other hand, a class-rebalancing data selection strategy is designed to fairly differentiate clean labels from noisy labels. Semi-supervised learning is performed accordingly by discarding noisy labels. Extensive experiments demonstrate the superiority of the proposed ERAT framework.

\end{abstract}


\IEEEpeerreviewmaketitle

\section{Introduction}
\label{sec:intro}

Deep neural networks (DNNs) have shown exceptional success in a wide range of scenarios, from image recognition to natural language processing \cite{zhang2021aggregation,tang2021frame,ouyang2021suppressing,chen2022gsmflow}. However, as the foundation, the availability of reliable training data is often a challenge in real-world applications, due to a variety of prevalent issues such as malicious attacks and inevitable errors \cite{yuan2021automa,li2023trustable,zhuo2022uncertainty}. Various attack mechanisms exist in real-life applications that intentionally poison training data with imperceptible perturbations to disrupt model training and steer it towards undesired outcomes, known as \textit{data poisoning attack}. DNNs show high vulnerability to data corruption by poisoning \cite{biggio2018wild,feng2019learning,huang2020unlearnable,fowl2021adversarial,yu2022availability}. There is also unintentional noise that occurs on labels due to inexperience or errors of annotators, as well as possible ambiguities of data. Owing to these, the annotation process is not always reliable and often incurs label noise
\cite{sukhbaatar2015training,ghosh2017robust,wang2019symmetric}. DNNs have considerable power to even memorize label noise patterns during the training process, resulting in inferior performance \cite{zhang2021understanding,tanaka2018joint,arpit2017closer}. These factors, i.e., data perturbations and label noise, pose significant threats to the integrity and reliability of DNNs, severely impacting their applicability, particularly in safety-critical applications. Therefore, it is of paramount importance to develop effective defense methods against deceptive corruptions on data and labels.

The detrimental effects of data and label corruptions caused by the noises discussed above have garnered much attention. Many attempts to overcome these problems have been made in response. One such example is adversarial training, which has been demonstrated to be an effective strategy for combating data corruption through perturbations \cite{tao2021better,huang2020unlearnable}. This technique is used to generate a series of poisoned examples by taking the model to be learned as the attacking target, whilst simultaneously learning to give accurate predictions on the poisoned examples. Some data transformation techniques \cite{liu2023image,qin2023learning} are also shown effective in defending against data corruption. In consideration of defense against label noise, a wide range of studies have also been conducted, spanning from modifying model structures to redesigning loss functions. Specifically, many methods are dedicated to offsetting the noise transition by designing noise adaptation blocks \cite{sukhbaatar2015training,goldberger2016training} or estimating a noise transition matrix \cite{patrini2017making,ma2018dimensionality}. Others endeavor to penalize noisy labels by re-weighting data \cite{ren2018learning,jiang2018mentornet}, removing potential wrong labels \cite{li2020dividemix,karim2022unicon}, or dropping mislabelled examples \cite{han2018co,nguyen2019self,shen2019learning,yu2019does,cheng2020learning,zhang2022ideal}.  Regularization techniques, such as dropout \cite{srivastava2014dropout}, ensembling \cite{laine2016temporal,tarvainen2017mean}, and data augmentation \cite{zhang2017mixup}, are commonly adopted to help penalize overfitting to noisy labels. While these methods showcase the potential to prevent noise influence to some extent, they overlook the co-existence of both data and label corruptions, hindering model effectiveness and applicability in real-world scenarios. MSPL \cite{zhang2023multi} is designed to address the two types of corruption simultaneously. However, it relies on the assumption that the corruption type is known in advance.

In this work, we aim to answer the question of whether it is possible to defend against data and label corruptions simultaneously in order to obtain a higher level robustness of models, where an Effective and Robust Adversarial Training (ERAT) method is proposed without prior knowledge regarding noise details. The ERAT strengthens model resilience through adversarial learning surrounding imaginary malicious data perturbations and semi-supervised learning based on noisy label identification and removal. In more detail, adversarial learning is performed between a perturbation generation module and the predictive model. The perturbation generation module produces multiple perturbations to poison data by taking the classification model as the target, while the model is trained to produce consistent and correct predictions on data of different views, i.e., augmentation and various perturbation views. To reduce the computational overhead and accelerate the training process with multiple perturbations, we leverage a hybrid adversarial training strategy by uniformly sampling different categories of perturbations in each training step. This enables the model to handle the case where the poisoning type is unknown, which increases model applicability. Furthermore, we identify and remove corrupted labels to guarantee that the model will be trained with reliable annotations. To achieve this, a novel operation is introduced to measure the extent to which the model agrees with the given label and disagrees with other labels. Based on the calculated scores, a class-rebalancing sample selection strategy is proposed to effectively remove potential noisy labels and divide the dataset into a labeled set and an unlabeled set. Semi-supervised learning is performed accordingly to train the model. Extensive experiments are conducted under various corruption settings, demonstrating the impressive performance of the proposed method in terms of effectiveness and robustness.

The main contributions are summarized as follows:
\begin{itemize}
\item We identify the negative effects of the co-existence of data and label corruptions on DNNs, and the weakness of existing robust learning methods.

\item A novel Effective and Robust Adversarial Training (ERAT) method is proposed. ERAT alleviates the effects of data and label corruptions by performing a hybrid adversarial training on imaginary data poisoning as well as a semi-supervised learning based on class-rebalancing sample selection. A series of novel learning strategies including multiple data poisoning and uniform sampling, multi-view alignment, and novel data scoring are devised accordingly to facilitate learning.

\item Comprehensive experiments in various corruption environments are conducted, and the results show that the proposed method is capable of managing learning scenarios with severe corruption.

\end{itemize}

\section{Related work}

\subsection{Data Poisoning Attack}

Data corruption focused on in this work refers in particular to date poisoning, which aims to manipulate model predicting behaviors by intentionally modifying the training data with imperceivable perturbations. Extensive studies have revealed the fragile natures of deep learning models to the presence of poisoned data \cite{koh2017understanding}. Existing poisoning methods can be roughly divided into two categories, i.e., poisoning a few data instances and poisoning whole datasets. The former posts limited threats to deep models as it targets influencing a few examples \cite{shafahi2019poison,chen2017targeted,yao2019latent}. By contrast, infecting whole datasets can invalidate models on large-scale normal data. And this kind of attack cannot be avoided by identifying and removing a few poisoned training samples. Feng \textit{et al.} \cite{feng2019learning} present the first work to produce training-time adversarial data by solving a bi-level optimization problem. They aim to learn a perturbation generator so that the target model trained on data polluted by the generator would make wrong predictions on clean unseen data. Huang \textit{et al.} \cite{huang2020unlearnable} propose to make a target model learning nothing from data by injecting err-minimizing noise. Yu \textit{et al.} \cite{yu2022availability} show that it is feasible to leverage adversarial attack methods \cite{goodfellow2014explaining,liudelving2014,chen2017targeted} to perform poisoning attacks.

Effective defense has not received enough attention, with just a few works proposed. \cite{tao2021better} is a representative one, which leverages adversarial training to improve model robustness to data perturbations. In this learning framework, a bunch of worse-case data are generated by taking the model as the victim and then using the model to enhance model resilience. The effectiveness of adversarial training against data corruption is also verified in \cite{huang2020unlearnable}. However, current adversarial training methods are developed under the assumption that the poison type is known. Some data transformations can also contribute to combating data corruption. For example, Image Shortcut Squeezing (ISS) \cite{liu2023image} utilizes image compression techniques to eliminate the negative influence of corruption. UErase \cite{qin2023learning} searches an error-maximizing adversarial augmentation from a set of augmentation policies in each iteration to help wipe out the negative effects of data corruption.

\begin{figure*}
\center
  \includegraphics[width=0.8\textwidth]{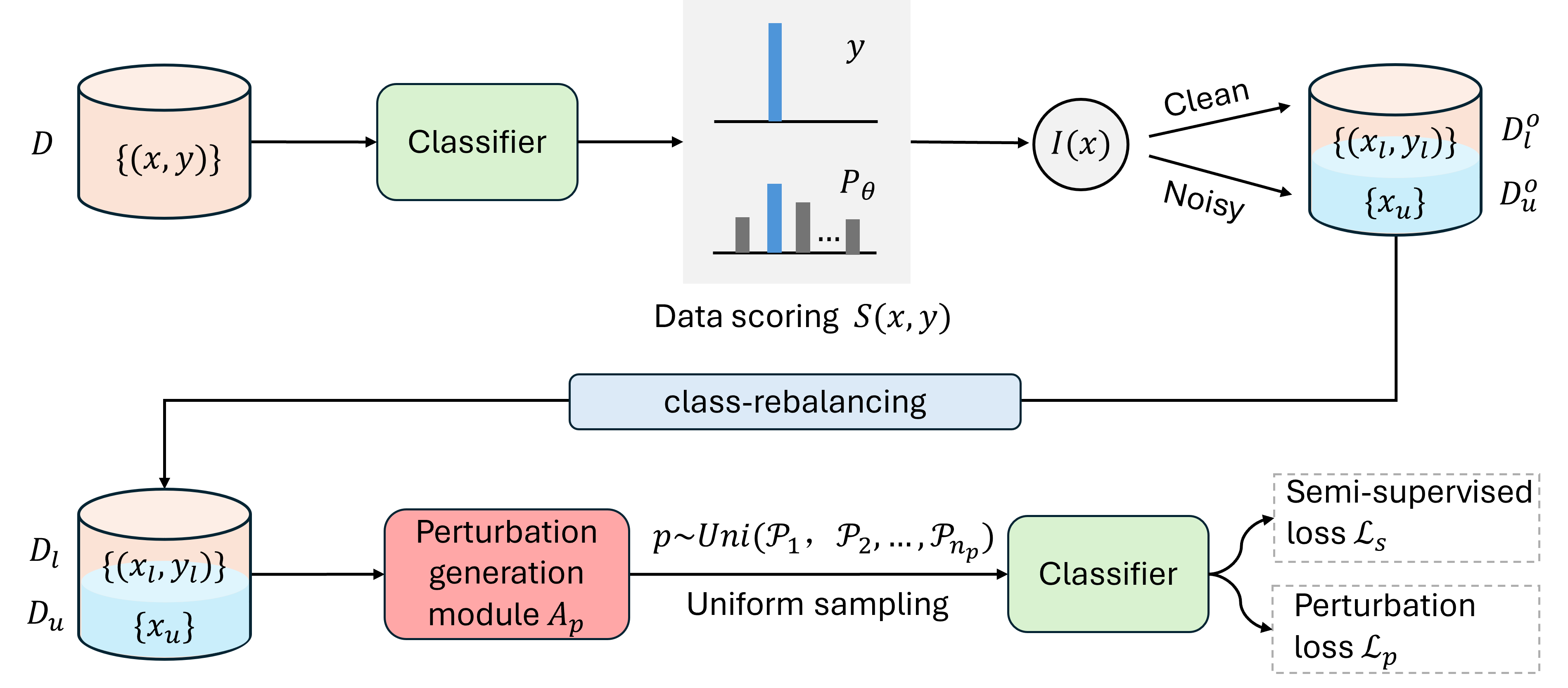}
   \caption{An illustration of the ERAT framework. In each epoch, the proposed method first leverages a scoring-based class-rebalancing strategy to separate the original dataset into a labeled set and an unlabeled set to include data with clean labels and noisy labels, respectively. Next, hybrid adversarial training is performed between the perturbation generation module and the classifier. The perturbation generation module uniformly samples attacking models to produce the most vicious data perturbations by enlarging the semantic gap between the perturbed data and the original data. The classifer is trained to maintain semantic consistency between original data of different augmentation and perturbation views by semi-supervised learning.}
  \label{fig1}
\end{figure*}

\subsection{Learning with Noisy Labels}

The existence of noisy labels can adversely influence the performance of machine learning models. To solve this, a line of work aims to measure noise patterns. For instance, Sukhbaatar \textit{et al.} \cite{sukhbaatar2015training,goldberger2016training} design a noise adaptation layer and add it on the top of the original model to mimic the transform from true labels to noisy labels. Dedicated networks \cite{han2018masking,goodfellow2014generative} are also designed to further handle more complex noise. Instead of modifying the model structure, calculating the label transition matrix to adjust the loss function is also a popular strategy. For example, Patrini \textit{et al.} \cite{patrini2017making} learn the transition matrix by means of an anchor point set. Xia \textit{et al.} \cite{xia2019anchor} propose to construct a transition matrix without the help of anchor points and update it with only a noisy validation set. Another strategy is to detect potential noisy labels and accordingly penalize their influence, including data re-weighting, wrong label removal and noisy example discarding. For instance, Ren \textit{et al.} \cite{ren2018learning} attempt to assign weights to data instances by measuring their importance based on the gradient direction. Jiang \textit{et al.} \cite{jiang2018mentornet} present a data-driven curriculum learning strategy, where a mentor network provides time-varying weights to the student network based on the feedback from the student network. Li \textit{et al.} \cite{li2020dividemix} fit a two-component Gaussian Mixture Model to the sample distribution to identify clean and noisy labels. Semi-supervised learning is performed by removing noisy labels. Karim \textit{et al.} \cite{karim2022unicon} design a uniform sample selection strategy to pursue fair clean and noisy label separation among classes for semi-supervised learning, and leverage contrastive learning to encourage effective feature learning. Han \textit{et al.} \cite{han2018co} train two networks in a collaborative way, where each network selects data with small loss as potential clean data to train its peer network. Nguyen \textit{et al.} \cite{nguyen2019self} compare the running averages of predictions on data instances with given labels to identify clean and noisy labels, and train the Mean-Teacher model only with clean labels. Cheng \textit{et al.} \cite{cheng2020learning} sieve out low-confidence examples and utilize peer loss to train the model on the remaining clean data. There are also methods proposed to redesign the loss function to minimize risk on data with original clean labels regardless of the existence of noisy labels. Representative work includes generalized cross entropy (GCE) \cite{zhang2018generalized}, symmetric cross entropy (SCE) \cite{wang2019symmetric}, mean absolute error (MAE) loss \cite{zhang2018generalized}, peer loss \cite{liu2020peer}, etc. Regularization is also taken as an effective tip to prevent overfitting to noise, e.g., dropout \cite{srivastava2014dropout}, ensembling \cite{laine2016temporal,tarvainen2017mean}, data augmentation \cite{zhang2017mixup} and so on.

While these methods have made progress, they commonly consider one type of corruption and overlook the truth of the concurrence of two kinds of corruption. Zhang \textit{et al.}  \cite{zhang2023multi} designs a Siamese prototype learning framework, where the adversarial training with inter- and intra-class distribution-based sample selections are leveraged to handle data and label corruptions. However, this method assumes that the corruption type is known. This significantly limits its real-world applications. In this work, we propose to realize robust learning against the co-existence of various corruptions without knowing the corruption type.

\section{Methodology}
\subsection{Preliminaries}
The problem of learning with corruptions is set up in the context of typical $K$-class classification, where the purpose is to learn a mapping function from an input space into a label space. Formally, denote a clean labeled dataset as $\mathcal{D}^c = \{({x}^c, {y}^c): {x}^c \in \mathcal{X}^c, {y}^c \in \mathcal{Y}^c\}$. ${{x}^c} \in \mathcal{X}^c \subseteq \mathbb{R}^{d}$ represents clean data, while ${{y}^c} \in \mathcal{Y}^c \subseteq \{0,1\}^K$ represents the associated true label. However, in real-world scenarios, obtaining such a perfect dataset may not be feasible, and instead, we can only observe a noisy dataset, denoted as ${\mathcal{D}} = \{({x}, {y}): {x} \in \mathcal{{X}}, {y} \in \mathcal{{Y}}\}$ for brevity. And our goal is to learn a robustness mapping function, which is able to associate original clean data with true labels, i.e., $f_\theta(\cdot): \mathcal{X}^c \rightarrow \mathcal{Y}^c$ with parameters $\theta$, when training with ${\mathcal{D}}$. Thus, during testing, the model can generalize well on unseen clean data. 

\subsection{Framework Overview}

To defend against dual corruption in the dataset, we propose a novel Effective and Robust Adversarial Training (ERAT) method. An illustration of the proposed method can be found in Figure \ref{fig1}. In each epoch, the ERAT first measures the difference between model predictions and given labels with the proposed scoring function. Class-rebalancing sample selection is then leveraged to divide the original dataset into a labelled set and an unlabeled set, which includes data with potential clean labels and noisy labels, respectively. With the separation result, hybrid adversarial training is performed between a perturbation generation module and the classification model. The perturbation generation module uniformly learns multiple types of perturbations by taking the classification model as the victim model. The classification model, in turn, learns to give consistent and correct predictions between original data and perturbed data with the semi-supervised learning strategy. 

\begin{figure}[tbh]
\begin{subfigure}[Clean label selection after warmup]
{\includegraphics[angle=0, width=0.23\textwidth]{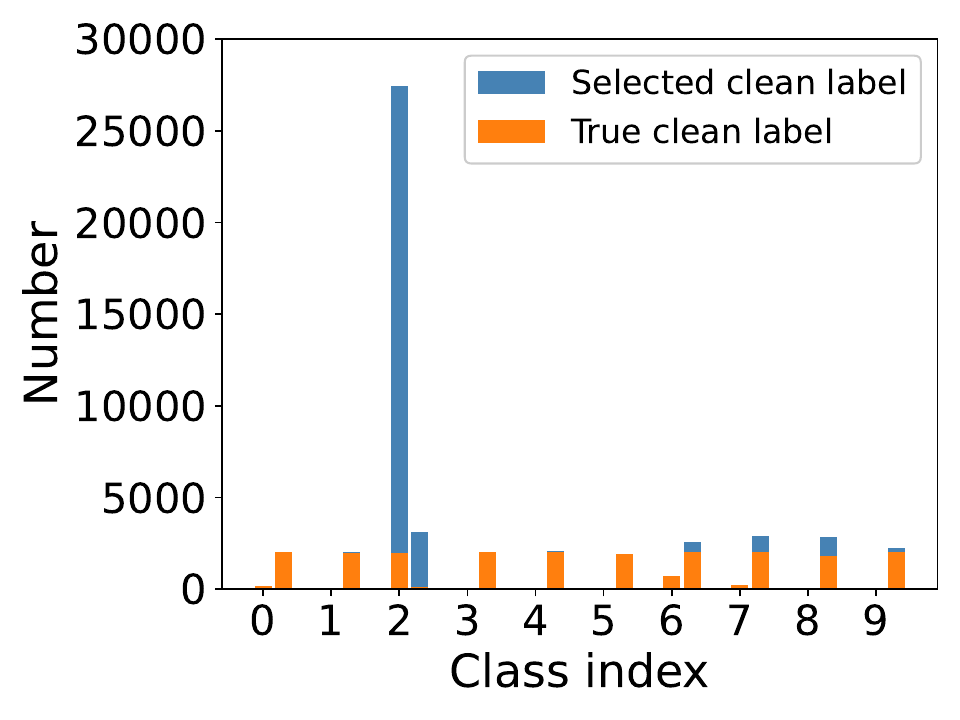}}
\end{subfigure}
\begin{subfigure}[Clean label selection after 60 epochs]
{\includegraphics[angle=0, width=0.23\textwidth]{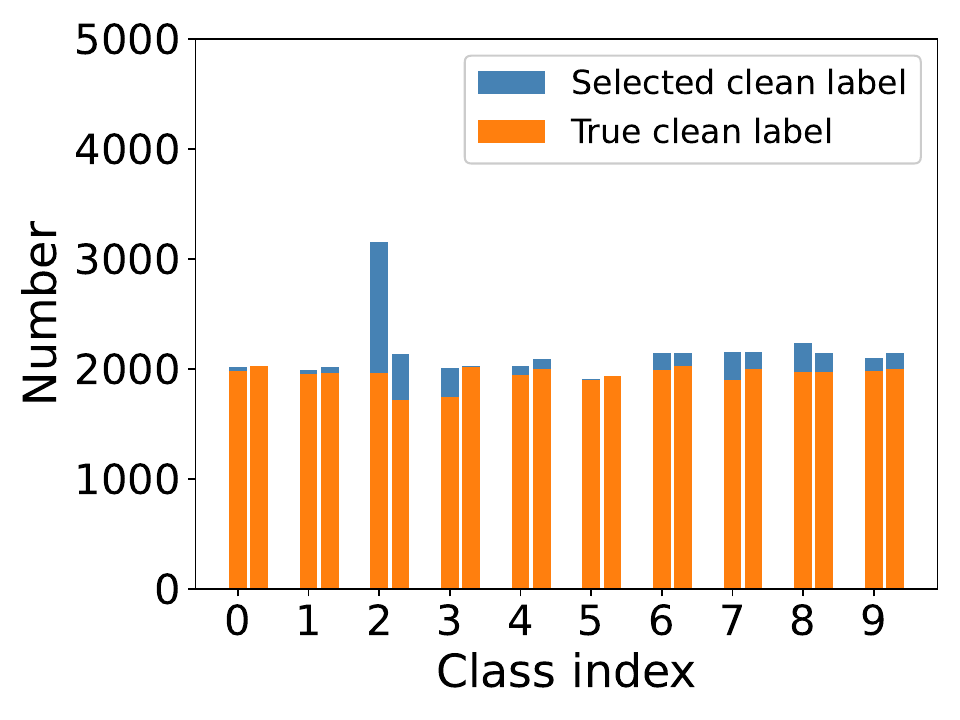}}\\
\end{subfigure}
\caption{A case of non-rebalancing and balancing selection for CIFAR-10 under $60\%$ instance-level label noise \cite{song2022learning}. The left and right bars in each class index represent non-rebalancing and rebalancing selection results, respectively. It can be seen that without rebalancing, the model has a strong bias towards the majority class, while ignoring other classes. The rebalancing strategy can effectively rectify this issue}
\label{pro}
\end{figure}

\subsection{Scoring-based Class-rebalancing Sample Selection}

Previous studies have shown that models tend to learn simple patterns before progressing to intricate ones during the learning process \cite{jiang2018mentornet,arazo2019unsupervised,han2018co,li2020dividemix,karim2022unicon}. In the scope of the proposed learning task, easy patterns would include samples with clean labels, samples of easy classes, and majority classes (mainly caused by label noise). In light of this, on the one hand, it is feasible to use model predictions at the early training stage to distinguish clean labels from noisy labels, where the model would agree more with clean labels than noisy labels \cite{jiang2018mentornet,arazo2019unsupervised}. We can follow the small-loss regime \cite{li2020dividemix,karim2022unicon} to measure the difference between predictions and given labels for differentiation. However, on the other hand, there might be an imbalance issue as the model would be prone to fit into easy classes or majority classes and thus produce high-confidence predictions on samples in these classes. Simply deeming high-confidence samples belonging to clean labels would result in an imbalanced labeled set, deteriorating the model quality \cite{li2020dividemix,karim2022unicon}. Considering this, we propose a class-rebalancing sample selection strategy based on a new scoring strategy to pursue an unbiased separation.

More specifically, given the classification model $f_\theta$, a novel scoring operation is designed to measure how much the model agrees with the given labels and disagrees with other labels:
\begin{equation}\label{eq1-2}
    \begin{split}
    &\mathcal{S}(x, {y}) = \|P_\theta(x) -  y\|^2_2,\\
   \end{split}
\end{equation}
where $P_\theta(\cdot)$ is the model prediction, i.e., the softmax probability.

Next, we use the score for clean and noisy label identification. Intuitively, a smaller score indicates that the model agrees more with the given label than other labels, so that the given label is more likely to be clean, or vice versa. Motivated by this, we propose to select samples with smaller scores from the original dataset to construct a labelled dataset to include potential clean labels, while the remaining are used to construct an unlabeled dataset by removing noisy labels. To achieve this, we set a threshold $\mu$ that can reflect the current training-set scoring distribution, and compare the score of each sample with the threshold to distinguish clean labels from noisy labels:
\begin{equation}\label{eq1-3}
    \begin{split}
    & \mathcal{I} (x) =\left\{\begin{array}{cl}{0} & {\textit{if} \ \ \mathcal{S}(x, {y}) > \mu} \\ {1} & {\textit{if} \ \  \mathcal{S}(x, {y}) \leq \mu,}\end{array}\right.\\
   \end{split}
\end{equation}
where $\mu = \frac{1}{|{\mathcal{D}}|}\sum_{x \in {\mathcal{D}}} \mathcal{S}(x, {y})$ is the average score over all samples in the dataset.

With the indicator, we can separate the original data into a labeled set $\mathcal{D}^o_l = \{({x}, {y}): \mathcal{I} (x) =1,  x \in \mathcal{D}\}$, and unlabeled sets $\mathcal{D}^o_u = \{({x}, {y}):\mathcal{I} (x) =0,  x \in \mathcal{D}\}$. However, small-score samples would appear more often in easy classes or majority classes, incurring the imbalanced selection. To alleviate it, we propose to rearrange the separation result, ensuring neglected classes to be selected more frequently.

The proposed strategy reassigns the number of clean labels in each class by additionally considering class-level difference instead of only instance-level difference. Different from the instance-level difference that is measured on the whole dataset, the class-level difference calculation is only based on the labeled dataset from the last epoch (In the first epoch, the labeled dataset is initialized with the original training dataset). The rationale behind this is that, after training with potential clean labels, the more confident the model is in a class, the more labels in this class would be selected as clean labels. Particularly, given the labelled set from the last epoch denoted as $\mathcal{D}^{\prime}_l$, the score for the class $k$ can be calculated as follows:
\begin{equation}\label{eq1-2}
    \begin{split}
    &\mathcal{S}_k = \frac{1}{|\mathcal{D}^{\prime}_{l,k}|} \sum_{(x,y) \in \mathcal{D}^{\prime}_{l,k}} \mathcal{S}(x, {y}).\\
   \end{split}
\end{equation} 

With the class scores, the sampling rate for the class $k$ can be defined as follows:
\begin{equation}\label{eq1-3}
    \begin{split}
    &\mathcal{R}_k = \frac{(1 - \mathcal{S}_k)^\eta}{\sum_{k=1}^{K} (1 - \mathcal{S}_k)^\eta},\\
   \end{split}
\end{equation}
where $\eta \in [0,1]$ is to tune the sampling rate.

With the ratio, we can choose the smallest-score samples in the class $k$ to construct the labeled dataset with the number of:
\begin{equation}\label{eq1-5}
    \begin{split}
    & d_k = |\mathcal{D}^o_l|  * \mathcal{R}_k.\\
   \end{split}
\end{equation}

The proposed rebalancing strategy still follows the small-loss selection principle. The difference from the instance-level selection is that the proposed method further considers neglected classes. This can avoid biased sample selection, promoting a high-quality model. An example of non-rebalancing and rebalancing selection can be found in Figure \ref{pro}.  After label corruption, the number of samples in each class is $[2031, 2015, 27634, 2026, 2091, 1937$, $2574, 2894, 4542, 2256]$, which is extremely imbalanced. After warmup with this dataset (detailed in Section \ref{sec:opt}), most of the smallest-score samples are concentrated in the majority class, i.e., class 2, and are selected as clean by the non-rebalancing method. We can also observe that after 60 epochs of training with rebalanced labelled data, the non-rebalancing selection is still biased towards some classes. Training with such an imbalanced and less accurate set would significantly influence performance. On the contrary, the proposed method can achieve high-quality sample selection, benefiting final results.

\subsection{Imaginary Data Poisoning}\label{lf}

The core idea of our work on defending against corruption by data poisoning is to continually generate malicious data perturbations to enhance model generalization and resilience. By enforcing the model to preserve semantic consistency between original data and perturbed data, we can make the model generalize well into unseen clean data. Inspired by the previous method \cite{tao2021better}, we propose to take advantage of adversarial training, where a perturbation generation module and the classification model fight against each other. The perturbation generation module endeavors to learn the most vicious data perturbations to attack the classification model, while the classification model is updated to give correct and consistent predictions regardless of the effects of any perturbation. Through the game theoretical framework, each party encourages effective update of each other, benefiting final results. 

Specifically, the adversarial perturbation $\delta$ for poisoning data is generated by the following objective:
\begin{equation}\label{eq1-1}
    \begin{split}
    & \min _{\{\delta\} \in \Delta} \mathcal{L}_{p} = - \mathbb{E}_{x \sim {\mathcal{D}}} [\|P_\theta(x+\delta) -  P_\theta(x)\|^2_2 + \ell(P_\theta(x+\delta), {y})],\\
   \end{split}
\end{equation}
where $\Delta$ is a constraint set. The objective aims to produce adversarial perturbations that can significantly change the semantics of the original data in both the feature space and label space. As the type of perturbation is usually unseen and training against a single type of perturbation cannot handle other types (an example of single-defense method AT can be seen in Table \ref{at}), we propose to generate multiple perturbations to ensure comprehensive defense. Nevertheless, training the model with multiple perturbations for each data instance simultaneously would lead to significant computational overhead.

To alleviate this, instead of using all perturbations for each data, we propose a hybrid adversarial training strategy by uniformly sampling each kind of perturbation from the perturbation set to poison data in each training step. In particular, denote the final rebalanced datasets as $\mathcal{D}_l$ and $\mathcal{D}_u$. Let $\mathcal{A}_p$ represent a poisoning attack model which generates one type of perturbation. The uniform poisoning process can be defined as:
\begin{equation}\label{eq1-8}
    \begin{split}
    &{x}^{adv}_l = \mathcal{A}_p({x}_l),\\
     &{x}^{adv}_u = \mathcal{A}_p({x}_u),\\
     &s.t. \ p \sim Uni (\{\mathcal{P}_1,\mathcal{P}_2, ..., \mathcal{P}_{n_p}\}),\\
     & \ \ \ \ \ {x}_l \sim Uni (\{{x}_l\}),\\
     & \ \ \ \ \ {x}_u \sim Uni (\{{x}_u\}),\\
   \end{split}
\end{equation}
where $Uni(\cdot)$ is a uniform sampling operation. In this work, we consider the $\ell_{p}-$ norm attack family, i.e., $\|\delta\|_{p} \leq \epsilon_{p}$, which is standard practice in data poisoning. $\epsilon_p$ is the perturbation magnitude, which is usually small to ensure the utility of modified data. The uniform data poisoning would create poisoning datasets denoted as $\mathcal{{D}}^{adv}_l$ and $\mathcal{{D}}^{adv}_u$.

\begin{table}[tb]
\center
\caption{Test accuracy $(\%)$ of AT\cite{tao2021better} with $l_{\infty}$ or $l_{2}$-norm defense against data poisoning on CIFAR-10 (without label noise).}\label{at}
\begin{tabular}{ccc cccc cccc cccc cccc}\toprule[1pt]

\multirow{1}{*}{{Method}} &\multirow{1}{*}{\emph{L2C$_{\infty}$}} &\multirow{1}{*}{\emph{UP$_{\infty}$}} &\multirow{1}{*}{\emph{AP$_2$}}&\multirow{1}{*}{\emph{UAP$_2$}} &\multirow{1}{*}{\emph{UHP$_2$}}&\multirow{1}{*}{\emph{URP$_2$}} \\\hline

\multicolumn{1}{c}{AT$_{\infty}$}    &{78.63}   &{75.26}  &{10.01} &{31.04}  &{21.16}   &{59.62} \\

\multicolumn{1}{c}{AT$_2$}     &{41.12}   &{13.27}  &{77.2} &{77.91}  &{81.05}   &{78.9} \\\hline
\toprule[1pt]
\end{tabular}
\end{table}

\subsection{Semi-supervised Learning with Multiple Poisoning Robustness}\label{f3} 

We train the model to keep semantic consistency between poisoned data and original data to improve model robustness and generalization abilities. To this end, we create two augmented versions for each sample, i.e., weak ($\mathcal{T}_w$) and strong augmentation ($\mathcal{T}_s$). For labelled data, we leverage the given labels for supervision. For unlabeled data, we use model predictions on weakly augmented data as Pseudo-labels. The final objective can be defined as follows:
\begin{equation}\label{eq1-9}
    \begin{split}
     &\min _{\theta} \mathcal{L}_s =  \mathcal{L}_l + \lambda \mathcal{L}_u\\
     & \mathcal{L}_l = \underset{({x}_l, {y}_l) \sim \mathcal{D}_l, \atop ({x}^{adv}_l, {y}_l) \sim \mathcal{{D}}^{adv}_{l}} {\mathbb{E}} [\ell(P_\theta(\mathcal{T}_s(x_l)), {y}) + \ell(P_\theta({x}^{adv}_l), {y}) \\
     & \ \ \ \ \ \ \ \ \ \ \ \ \ \ \ \ \ \ \ \ \ \ \ \ \ \ \ \ + \ell(P_\theta(\mathcal{T}_s({x}^{adv}_l)), {y})],\\
     & \mathcal{L}_u = \underset{{x}_u \sim \mathcal{D}_u,  \atop {x}^{adv}_u \sim \mathcal{{D}}^{adv}_u}  {\mathbb{E}} [\|P_\theta(\mathcal{T}_s (x_u))  -  {P}_{\tilde{\theta}}(\mathcal{T}_w (x_u))\|^2_2 \\
     & \ \ \ \ \ \ \ \ \ \ \  + \|P_\theta({x}^{adv}_u) -  {P}_{\tilde{\theta}}(\mathcal{T}_w(x_u))\|^2_2]\\
     &  \ \ \ \ \ \ \ \ \ \ \  + \|P_\theta(\mathcal{T}_s({x}^{adv}_u)) -  {P}_{\tilde{\theta}}(\mathcal{T}_w (x_u))\|^2_2],\\
   \end{split}
\end{equation}
where $\lambda > 0$ is a balancing parameter. `` $\tilde{\cdot}$ '' means that model parameters are frozen. 

\subsection{Optimization Algorithm}
\label{sec:opt}

\begin{algorithm}[t!]
       \caption{Effective and Robust Adversarial Training}
       \label{alg1}
       \textbf{Require}: Randomly initialized network $f_\theta$, training data $\mathcal{D} = \{({x}, {y}): {x} \in \mathcal{X}, {y} \in \mathcal{Y}\}$, mini-batch size $m$, step size $\pi$, perturbation budget $\epsilon$, parameters $\eta, \lambda$\;
        \textbf{Output}: Networks $f_\theta$;\\
       $f_\theta = warmup (f_\theta,\mathcal{A}_p,\mathcal{D})$ // Warmup the network;\\
        // Formal Training;\\
        $\mathcal{D}_l = \mathcal{D}$// Initialize the labelled dataset;\\
        \Repeat{Maximum iteration times}
        {
        $\mathcal{D}^o_l, \mathcal{D}^o_u \leftarrow \mathcal{I}(\mathcal{S}(x, {y}))$ // Score samples to obtain raw sample selection;\\ 
        $\mathcal{S}_k, \mathcal{R}_k \leftarrow  \{\mathcal{S}(x, {y}), \{(x, {y})\} \in \mathcal{D}_l: k= 1 ..,K\}$ // Obtain the class-level score and sampling rate;\\ 
        $d_k = |\mathcal{D}^o_l| *\mathcal{R}_k, k= 1 ..,K$// Re-balance selection;\\
        $Id_k \leftarrow \{argsort(\mathcal{S}^k(x, {y})), d_k : k= 1 ..,K\}$ // Sort data score in each class to obtain the index of $d_k$ smallest-score value;\\
        $ \mathcal{{D}}_{l} \leftarrow \{({x}_i, {y}_i): {i} \in Id_k\}, \mathcal{{D}}_{u} \leftarrow \mathcal{{D}} \backslash \mathcal{D}_{l}$ // Obtain the clean and noisy sets;\\
        \Repeat{Maximum iteration times}
        {
        $\{({x}_l, {y}_l)\},\{{x}_u\} \sim \mathcal{{D}}_{l},\mathcal{{D}}_{u}$ //  Sample mini-batch;\\
        \For{$i = 1 \rightarrow n_p$}{
            $p \sim Uni (\{\mathcal{P}_1,\mathcal{P}_2, ..., \mathcal{P}_{n_p}\})$// Uniformly sample the poisoning type;\\
            ${x}_l, {x}_u \sim Uni (\{{x}_l\}), Uni (\{{x}_u\}$// Uniformly sample labeled and unlabelled data;\\
        \For{$epoch = 1 \rightarrow T$}{
            ${\delta}_l^{t+1}/{\delta}_u^{t+1}= Clip (({\delta}_l^t/{\delta}_u^{t+1}-\pi \cdot \operatorname{sign}(\nabla_{{x}_l/{x}_u} \mathcal{L}_{p})),-\epsilon,\epsilon )$  // Update adversarial perturbations;\\ 
        }
        ${x}^{adv}_l/{x}^{adv}_u \leftarrow Clip (x_l/x_u +{\delta}_l/{\delta}_u, 0, 1)$// Generate adversarial labeled mini-batch;\\
        }
        $\theta \leftarrow {\theta} - \alpha \cdot \nabla_{\theta} \mathcal{L}_s$// Update the model;\\
        }
        }
\end{algorithm}

The optimization scheme is summarized in \textbf{Algorithm \ref{alg1}}. Before performing our proposed adversarial training, we would first warmup the network with the conventional CE loss: 
\begin{equation}\label{eq2-1}
    \begin{split}
     & \min _{\theta} \mathcal{L} = \mathbb{E}_{({x}, {y}) \sim \mathcal{D}} [\ell(P_\theta(\mathcal{T}_s({x})), {y}) \\
     &  \ \ \ \ \ \ \ \ \ \ \ \ \ \ \ \ \ \ \  +  \ell(P_\theta(\mathcal{T}_s({x}^{adv})), {y})].\\
   \end{split}
\end{equation}

To learn adversarial perturbations, we adopt the projected gradient descent (PGD) method:
\begin{equation}\label{eq2-2}
    \begin{split}
     & {\delta}^{t+1}= Clip \left(\left({\delta}^t-\pi \cdot \operatorname{sign}\left(\nabla_{{x}} \mathcal{L}_{p} \right)\right),-\epsilon,\epsilon \right),\\
     &s.t. \ {\delta}^{0}= 0,\\
   \end{split}
\end{equation}
where $\pi$ denotes the step size. $t$ is the current perturbation step with total steps of $T$. $Clip(\cdot,{-\epsilon},{\epsilon})$ restricts the perturbation to be within a ${\epsilon}-$ball around the input $x$. At the end of each generation, we would also project the generated adversarial data into a $[0,1]$ space, i.e., ${x}^{adv} = Clip(x+\delta,0,1)$. And the final perturbation would be $\delta = {x}^{adv} -x$.

\section{Experiments}

The effectiveness of the ERAT is verified by conducting extensive experiments to answer the following research questions (RQ):
\begin{enumerate}[leftmargin = 16pt]
 \item Do existing robust learning methods still perform well when facing the existence of data and label corruptions? How good performance is achieved by the ERAT over existing methods under different corruption settings?
  \item How accurately does the proposed method perform under various noise magnitudes? What budgets does the proposed method require for effective defense?
 \item Does the proposed method perform well on different networks?
  \item How does each component of the proposed method help improve the robustness?
  \item How does the efficiency of the proposed method compare to the strategy that conducts respective training when dealing with multiple perturbations over the training dataset?

\end{enumerate}

\subsection{Datasets}

Three widely-used datasets for supervised learning with noisy labels and data perturbations are selected, i.e., CIFAR-10, CIFAR-100 \cite{krizhevsky2009learning}, and Tiny-ImageNet \cite{le2015tiny}. CIFAR-10 and CIFAR-100 contain 60,000 images with size 32 × 32, categorized into 10 and 100 concepts, respectively. In the experiments, each dataset is split into a training set comprising 50,000 data instances and a test set consisting of 10,000 data instances. Tiny-ImageNet is a subset of ImageNet, consisting of 64 × 64 images of 200 classes. Each class contains 500 training images and 50 test images.

To comprehensively verify the robust learning ability of the proposed method, we poison each dataset with different combinations of data and label corruptions.

\noindent\textbf{Data corruption.} Six representative data corruption methods are selected to poison datasets:
\begin{enumerate}[leftmargin = 16pt]
  \item \textbf{Training-time adversarial perturbations (L2C)} \cite{feng2019learning}. This method learns a generative model to produce adversarial perturbations by solving a bi-level optimization problem:
      \begin{equation}\label{eq3-1}
        \centering
        \begin{split}
        & \max _{\xi} \mathbb{E}_{(x, y) \sim \mathcal{D}} [\ell(f_{\theta^{*}{(\xi)}}(x), {y})],\\
        & s.t. \ \ \theta^{*}{(\xi)}=\arg \min _{\theta} \mathbb{E}_{(x, y) \sim \mathcal{D}} [\ell(f_\theta(x + {q_\xi}(x,\epsilon)),y)],\\
        \end{split}
    \end{equation}
    where $q_\xi$ is the perturbation generator ($\xi$ is model parameters). In this work, we use L2C to generate \textit{sample-wise} perturbations for corruption; 
  \item \textbf{Unlearnable perturbations (UP)} \cite{huang2020unlearnable}. The UP learns to produce error-minimizing adversarial examples: $\min _{\delta}  \mathbb{E}_{(x, y) \sim \mathcal{D}} [ \min _{\theta} \ell(f_\theta(x + \delta),y)]$.
      In this work, we use L2C to generate \textit{class-wise} perturbations for corruption; 
  \item \textbf{AutoAttack (AA)} \cite{croce2020reliable}. It is a parameter-free PGD attack \cite{madry2018towards};
      
  \item \textbf{Adversarial perturbations (AP)} \cite{tao2021better}. The aim of the attack method is to mislead models to misidentify target data as a target class $t$: $\min _{\delta}  \mathbb{E}_{(x, y) \sim \mathcal{D}} [ \ell(f_{\theta^{\prime}}(x + \delta),t)]$;
  \item \textbf{Universal Adversarial perturbations (UAP)} \cite{tao2021better}. A universal version of AP that generates \textit{class-wise} perturbations;
  \item \textbf{Universal Hypocritical perturbations (UHP)} \cite{tao2021better}. This method seeks to produce \textit{class-wise} perturbations to help the model make correct predictions: $\min _{\delta}  \mathbb{E}_{(x, y) \sim \mathcal{D}} [ \ell(f_{\theta^{\prime}}(x + \delta),y)]$;
  \item \textbf{Universal random perturbations (URP)} \cite{tao2021better}. It uses the same random noise to poison the data of each class.
\end{enumerate}

\noindent\textbf{Label corruption.} The existing three kinds of label noise in the literature \cite{song2022learning} are all utilized, including:
\begin{enumerate}[leftmargin = 16pt]
  \item \textbf{Instance-dependent label noise (Inst.)}: following \cite{xia2020part}, samples are wrong annotated based on both true labels and features;
  \item \textbf{Symmetric label noise (Symm.)}: true labels are randomly flipped into other labels with a probability of $\kappa$;
  \item \textbf{Asymmetric label noise (Asymm.)}: samples are annotated by another similar class with a probability of $\kappa$.
\end{enumerate}

In addition to dual corruption, we also vary the corruption magnitude to create original experimental settings for compared methods. This ensures a fair comparison, further helping demonstrate the applicability of the proposed method.

\subsection{Setup}

\begin{table}[tb]
\center
\caption{Test accuracy $(\%)$ on clean data.}\label{Tcd}
\begin{tabular}{ccc cccc cccc cccc cccc}\toprule[1pt]

\multirow{1}{*}{{Method}} &\multirow{1}{*}{ CIFAR-10} &\multirow{1}{*}{CIFAR-100} &\multirow{1}{*}{Tiny-ImageNet} \\\hline

\multicolumn{1}{c|}{CE}  &{95.20}   &{66.48} &{57.4} \\

\multicolumn{1}{c|}{AT}   &{89.05}  &{62.35} &{51.62}\\

\multicolumn{1}{c|}{ISS}  &{86.35}   &{59.8}  &{49.74} \\

\multicolumn{1}{c|}{DAT}  &{84.2} &{60.1} &{52.9}  \\

\multicolumn{1}{c|}{DivideMix}  &{93.5} &{63.2} &{62.89}  \\

\multicolumn{1}{c|}{UNICON}  &{95.6} &{72.5}  &{63.1}  \\

\multicolumn{1}{c|}{MSPL} &{92.65}&{71.25}&{61.56}\\

\multicolumn{1}{c|}{ERAT}  &{93.78} &{69.93}  &{62.94} \\\hline
\toprule[1pt]
\end{tabular}
\end{table}

\begin{table}[tb]
\center
\caption{Test accuracy $(\%)$ on CIFAR-10. \textbf{Bold} indicates the best and the second best is \underline{underlined}.}\label{cifar10_acc}
\scalebox{0.8}{
\begin{tabular}{llc cccc cccc cccc cccc}\toprule[1.2pt]
\multirow{2}{*}{\emph{Corruption}} &\multirow{2}{*}{{Method}} &\multirow{1}{*}
{\emph{L2C}} &\multirow{1}{*}{\emph{UP}} &\multirow{1}{*}{\emph{AA}} &\multirow{1}{*}{\emph{AP}}&\multirow{1}{*}{\emph{UAP}} &\multirow{1}{*}{\emph{UHP}}&\multirow{1}{*}{\emph{URP}} \\
& &\multicolumn{3}{c}{\emph{$l_{\infty},\epsilon^{\prime} =8/255$}} &\multicolumn{4}{c}{\emph{$l_{2},\epsilon^{\prime} =0.5$}}\\ \toprule[1pt] 

\multirow{7}{*}{\tabincell{c}{\emph{Inst}. \\\emph{$\kappa = 0.6$}}}

&\multicolumn{1}{|c}{CE}  &{12.79}   &{10.89} &{{13.03}} &{20.31}     &{11.15}  &{9.17}  &{33.27}\\

&\multicolumn{1}{|c}{AT}   &{19.93}  &{9.96}  &{{10.01}} &{10.42}    &{20.16}  &{10.04}  &{23.38} \\

&\multicolumn{1}{|c}{ISS}  &{19.63}   &{10.0}  &{{22.97}} &{24.08}  &{10.53}   &{11.39}  &{12.61} \\

&\multicolumn{1}{|c}{DAT}  &{10.48} &{10.92} &{{22.36}} &{29.58}    &{9.86}  &{9.83} &{29.91}  \\

&\multicolumn{1}{|c}{DivideMix}  &{15.48} &{12.52} &{{38.22}} &{52.2}   &{11.13}  &{23.37} &{22.63}  \\

&\multicolumn{1}{|c}{UNICON}  &{44.04} &{22.32}  &{{59.7}} &{71.88}    &{18.9}  &{20.98}  &{39.25} \\

&\multicolumn{1}{|c}{MSPL}  &{\underline{{72.07}}} &{\underline{{61.78}}}  &{\underline{{66.56}}} &{\underline{{74.84}}}   &{{\bf{90.73}}}  &{{\bf{89.33}}}  &{\underline{{77.33}}} \\

&\multicolumn{1}{|c}{ERAT}  &\bf{83.71} &\bf{69.93}  &\bf{{85.66}} &\bf{86.1}    &\underline{88.94}  &\underline{86.16}  &\bf{89.15} \\\hline

\multirow{7}{*}{\tabincell{c}{\emph{Symm.} \\\emph{$\kappa = 0.6$}}}

&\multicolumn{1}{|c}{CE}   &{13.02} &{15.71}  &{{32.56}} &{37.21} &{11.85}  &{13.22}  &{39.85} \\

&\multicolumn{1}{|c}{AT}  &{{61.39}}  &{28.79}  &{{59.63}} &{61.94}    &{65.21}  &{66.02}  &{64.91}  \\

&\multicolumn{1}{|c}{ISS}  &{23.03}   &{24.91}  &{{26.8}} &{28.0}  &{11.68}   &{11.43}  &{16.33} \\

&\multicolumn{1}{|c}{DAT} &{19.63} &{9.62}  &{{53.37}} &{33.39}   &{12.87}  &{10.02} &{40.04}   \\

&\multicolumn{1}{|c}{DivideMix}  &{22.58}  &{16.21}  &{{63.03}} &{71.02}  &{16.54}  &{24.12} &{13.17}  \\

&\multicolumn{1}{|c}{UNICON}  &{45.92} &{23.21}  &{{72.88}} &{\underline{82.11}}    &{14.68}  &{20.02}  &{41.44} \\

&\multicolumn{1}{|c}{MSPL}  &{{\underline{78.31}}} &{{\underline{71.58}}}  &{\underline{{76.42}}} &{{{81.89}}}    &{{\underline{82.06}}}  &{{\underline{82.37}}}  &{{\underline{82.29}}} \\

&\multicolumn{1}{|c}{ERAT}  &\bf{85.04} &\bf{82.31}  &\bf{{85.51}} &\bf{85.84} &\bf{83.99}  &\bf{83.44}  &\bf{86.44}  \\\hline

\multirow{7}{*}{\tabincell{c}{\emph{Asymm.} \\\emph{$\kappa = 0.4$}}}

&\multicolumn{1}{|c}{CE}  &{15.99} &{13.44}  &{{33.97}} &{33.5}   &{10.0}  &{10.0}  &{12.72}  \\

&\multicolumn{1}{|c}{AT}  &{71.86}  &{\underline{69.65}}  &{{70.01}} &{80.9}    &{80.05}  &{81.88}  &{80.31}  \\

&\multicolumn{1}{|c}{ISS}  &{10.22}   &{10.09}  &{{10.34}} &{34.03}  &{10.08}   &{10.2}  &{16.83} \\

&\multicolumn{1}{|c}{DAT} &{13.43} &{11.21} &{{39.77}} &{52.53}   &{10.0}  &{10.01} &{53.71}   \\

&\multicolumn{1}{|c}{DivideMix}  &{22.54} &{11.78} &{{28.24}} &{62.82}   &{15.59}  &{13.25} &{13.18}  \\

&\multicolumn{1}{|c}{UNICON}  &{46.7} &{25.37}  &{{75.98}} &{82.24}    &{22.98}  &{24.45}  &{39.9} \\

&\multicolumn{1}{|c}{MSPL}  &{{\underline{77.67}}} &{{69.42}}  &{\underline{{80.59}}} &{\underline{{85.07}}}    &{{\underline{81.17}}}  &{{\underline{84.32}}}  &{{\bf{91.06}}} \\

&\multicolumn{1}{|c}{ERAT}  &\bf{85.87} &\bf{75.16}  &\bf{{87.99}} &\bf{90.63} &\bf{86.89}  &\bf{85.08}  &\underline{90.46}  \\
\toprule[1pt]
\end{tabular}
}
\end{table}
\noindent\textbf{Implementation details.} We take ResNet34 \cite{he2016deep} for the proposed method and all compared methods in basic experiments. We also test all methods with other backbones including VGG11, VGG16 \cite{simonyan2014very}, ResNet18, ResNet50 \cite{he2016deep}, WideResNet and ViT-B$/$16 \cite{dosovitskiy2020image} to further demonstrate the effectiveness of the proposed method. The momentum of the network is set as $0.9$, the weight decay is $5e^{-4}$, and the learning rate is $5e^{-2}$ with a decay of 0.1 at 60 epochs. The batch $m=64$. $\eta= 0.1, T = 7, \lambda= 25 \cdot clip(t_e/16, 0,1)$. $t_e$ is the training step. The training time is 100 for two datasets. On CIFAR-10, 10 epochs are spent to warmup the model and the remaining is used for formal training, while on CIFAR-100, we warmup the model via 30 training epochs, and formally train the model via 70 epochs. The weak augmentation is the random transformation of erasing and flipping. And we leverage AutoAugment \cite{cubuk2018autoaugment} to strongly augment data. The data poisoning attack model set used in training consists of $\ell_{\infty}$-norm and $\ell_2$-norm bounded models. 

\noindent\textbf{Baselines.} Since defending against different types of corruption remains unexploited, we consider the following baselines that focus on one type of corruption:
\begin{enumerate}[leftmargin = 16pt]
  \item \textbf{AT} \cite{tao2021better}, which performs adversarial training to train models with imaginary adversarial data to fight against poisoning attacks. It is noted that the method only trains the model on one type of adversarial examples given the sample type is known;
  \item \textbf{Image Shortcut Squeezing (ISS)} \cite{liu2023image}, which utilizes image compression techniques to eliminate the negative influence of data corruption;
  \item \textbf{DAT} \cite{qu2021dat}, which is developed to conduct robust learning with noisy labels by domain adaptation between clean labels and noisy labels;
  \item \textbf{DivideMix} \cite{li2020dividemix}, which leverages semi-supervised learning by identifying and discarding noisy labels via GMM on loss distribution;
  \item \textbf{UNICON} \cite{karim2022unicon}, which designs uniform sample selection to solve the class-imbalance problem when utilizing semi-supervised learning to train the model;
  \item \textbf{MSPL} \cite{zhang2023multi}, which leverages a Siamese prototype learning to improve model robustness. An adversarial training strategy with inter- and intra-class data selections is designed to handle data and label corruption with the information of the corruptions is known.

\end{enumerate}

We also report the results of the method \textbf{CE}, which is a very basic method that simply uses the CE loss to train the model.

\begin{table}[tb]
\center
\caption{Test accuracy $(\%)$ on CIFAR-100. \textbf{Bold} indicates the best and the second best is \underline{underlined}.}\label{cifar100_acc}
\scalebox{0.8}{
\begin{tabular}{llc cccc cccc cccc cccc}\toprule[1.2pt]
\multirow{2}{*}{\emph{Corruption}} &\multirow{2}{*}{{Method}} &\multirow{1}{*}
{\emph{L2C}} &\multirow{1}{*}{\emph{UP}} &\multirow{1}{*}{\emph{AA}} &\multirow{1}{*}{\emph{AP}}&\multirow{1}{*}{\emph{UAP}} &\multirow{1}{*}{\emph{UHP}}&\multirow{1}{*}{\emph{URP}} \\
& &\multicolumn{3}{c}{\emph{$l_{\infty},\epsilon^{\prime} =8/255$}} &\multicolumn{4}{c}{\emph{$l_{2},\epsilon^{\prime} =0.5$}}\\ \toprule[1pt] 

\multirow{7}{*}{\tabincell{c}{\emph{Inst}. \\\emph{$\kappa = 0.6$}}}

&\multicolumn{1}{|c}{CE}  &{3.84}   &{1.51}  &{{18.22}} &{16.21}  &{2.17}  &{2.42}  &{18.43} \\

&\multicolumn{1}{|c}{AT}  &{6.95}  &{5.82}   &{{7.28}} &{8.06}   &{7.57}  &{8.24}  &{7.41} \\

&\multicolumn{1}{|c}{ISS}  &{7.16}   &{4.43}  &{{9.81}} &{8.92}  &{0.99}   &{1.04}  &{2.11} \\

&\multicolumn{1}{|c}{DAT}  &{2.44} &{1.17} &{{17.02}} &{18.38}    &{1.09}  &{3.78} &{18.75}  \\

&\multicolumn{1}{|c}{DivideMix}  &{25.9} &{1.72}   &{{21.51}}  &{34.03}   &{4.49}  &{5.74}  &{5.47} \\

&\multicolumn{1}{|c}{UNICON} &{36.61} &{9.48}   &{{50.88}}  &{\underline{52.79}} &{18.32}  &{17.55}  &{41.89}  \\

&\multicolumn{1}{|c}{MSPL} &{\underline{{42.17}}} &{\bf{{38.78}}}   &\bf{{53.77}}  &{{49.28}} &{\underline{{46.42}}} &{\underline{{45.99}}}  &{\underline{{49.27}}}  \\

&\multicolumn{1}{|c}{ERAT}  &\bf{58.56} &\underline{38.32}   &\underline{{53.56}}  &\bf{55.18}    &\bf{55.73}  &\bf{56.94}  &\bf{54.15} \\\hline

\multirow{7}{*}{\tabincell{c}{\emph{Symm.} \\\emph{$\kappa = 0.6$}}}

&\multicolumn{1}{|c}{CE}  &{6.69}  &{1.77}   &{{20.05}} &{19.24}  &{12.62}  &{18.23}  &{20.51}  \\

&\multicolumn{1}{|c}{AT}  &{9.72}  &{7.93}   &{{10.02}} &{12.08}   &{10.88}  &{11.25}  &{10.54} \\

&\multicolumn{1}{|c}{ISS}  &{10.79}   &{8.88}   &{{9.59}} &{11.06}  &{1.56}   &{1.27}  &{9.21} \\

&\multicolumn{1}{|c}{DAT}  &{2.88}  &{1.58}   &{{19.32}}  &{20.01}   &{20.94}  &{21.28} &{21.63} \\

&\multicolumn{1}{|c}{DivideMix}  &{15.18} &{1.68} &{{32.64}}  &{49.77}   &{8.6}  &{8.35}  &{3.95}\\

&\multicolumn{1}{|c}{UNICON}   &{30.04}&{9.2} &{{51.0}}  &{52.73}   &{18.73}  &{19.34}  &{37.95} \\

&\multicolumn{1}{|c}{MSPL} &{\underline{{52.88}}} &{\bf{{44.58}}}   &\underline{{52.13}}  &{\bf{{59.07}}} &{\underline{{51.63}}}  &\bf{{58.81}}  &{\bf{{59.97}}}  \\

&\multicolumn{1}{|c}{ERAT}  &\bf{57.02} &\underline{36.82}   &\bf{{53.57}}  &\underline{54.61} &\bf{56.18}  &\underline{56.03}  &\underline{56.73}  \\\hline

\multirow{7}{*}{\tabincell{c}{\emph{Asymm.} \\\emph{$\kappa = 0.4$}}}

&\multicolumn{1}{|c}{CE}   &{10.55} &{1.31}   &{{34.92}}  &{31.74}  &{1.53}  &{24.41}  &{20.83}  \\

&\multicolumn{1}{|c}{AT} &{22.2} &{1.0}   &{{20.41}}  &{23.77}   &{23.34}  &{6.35}  &{21.05}  \\

&\multicolumn{1}{|c}{ISS}  &{30.5}   &{8.78}  &{{27.07}} &{18.95}  &{1.41}   &{1.21}  &{1.35} \\

&\multicolumn{1}{|c}{DAT}  &{5.52}  &{1.03}   &{{25.36}} &{36.2}    &{2.73}  &{2.36} &{40.76} \\

&\multicolumn{1}{|c}{DivideMix}  &{1.0} &{2.13}   &{{37.96}}  &{36.06}   &{6.74}  &{5.47}  &{2.14} \\

&\multicolumn{1}{|c}{UNICON}  &{48.19} &{8.24}  &{{42.91}}  &{\underline{50.14}}  &{13.83}  &{14.08}  &{18.99} \\

&\multicolumn{1}{|c}{MSPL} &{\underline{{49.46}}} &{\underline{{35.09}}}   &{\underline{{49.65}}}  &{{50.02}} &{\underline{{36.08}}}  &{\underline{{37.72}}}  &{\underline{{52.45}}}  \\

&\multicolumn{1}{|c}{ERAT}  &\bf{61.3} &\bf{43.62}   &\bf{{58.48}}  &\bf{61.44} &\bf{62.21}  &\bf{60.45}  &\bf{61.57}  \\
\toprule[1.2pt]
\end{tabular}
}
\end{table}

\begin{table}[tb]
\center
\caption{Test accuracy $(\%)$ on Tiny-ImageNet. \textbf{Bold} indicates the best and the second best is \underline{underlined}.}\label{Tiny_acc}
\scalebox{0.8}{
\begin{tabular}{llc cccc cccc cccc cccc}\toprule[1.2pt]
\multirow{2}{*}{\emph{Corruption}} &\multirow{2}{*}{{Method}} &\multirow{1}{*}{\emph{L2C}} &\multirow{1}{*}{\emph{UP}} &\multirow{1}{*}{\emph{AA}} &\multirow{1}{*}{\emph{AP}}&\multirow{1}{*}{\emph{UAP}} &\multirow{1}{*}{\emph{UHP}}&\multirow{1}{*}{\emph{URP}} \\
& &\multicolumn{6}{c}{\emph{$l_{\infty}, \epsilon^{\prime} =8/255$}} \\ \toprule[1pt] 

\multirow{7}{*}{\tabincell{c}{\emph{Inst}. \\\emph{$\kappa = 0.6$}}}

&\multicolumn{1}{|c}{CE}  &{5.89}   &{0.44}   &{{4.6}} &{4.81}  &{0.75}  &{1.06}  &{5.15} \\

&\multicolumn{1}{|c}{AT}  &{5.22}   &{0.48}   &{{4.69}} &{7.02}  &{5.64}   &{5.49}  &{3.54} \\

&\multicolumn{1}{|c}{ISS}  &{1.82}   &{0.65}   &{{2.21}} &{4.39}  &{0.57}   &{0.61}  &{1.74} \\

&\multicolumn{1}{|c}{DAT}  &{11.15} &{0.52}   &{{11.15}} &{11.83}    &{0.55}  &{0.39} &{1.87} \\

&\multicolumn{1}{|c}{DivideMix}  &{16.96}  &{0.4}  &{{13.14}}  &{11.86}   &{0.51}  &{0.59}  &{0.46} \\

&\multicolumn{1}{|c}{UNICON} &\bf{43.27} &{0.54}   &{{34.03}} &{37.66} &{0.99}  &{1.18}  &{2.23}  \\

&\multicolumn{1}{|c}{MSPL} &{{34.25}} &{\underline{{11.03}}}   &{\underline{{35.68}}}  &{\bf{{40.17}}} &{\underline{{26.08}}}  &{\underline{{23.32}}}  &{\underline{{38.65}}}  \\

&\multicolumn{1}{|c}{ERAT}  &\underline{39.37} &\bf{16.09}    &\bf{{38.54}}  &\underline{39.46} &\bf{26.72}    &\bf{29.28}  &\bf{41.41} \\ 
\hline

\multirow{7}{*}{\tabincell{c}{\emph{Symm.} \\\emph{$\kappa = 0.6$}}}

&\multicolumn{1}{|c}{CE}  &{4.93}  &{0.89}   &{{4.54}} &{4.59}  &{1.54}  &{2.14}  &{5.74}  \\

&\multicolumn{1}{|c}{AT}  &{7.47}  &{7.56}   &{{15.57}}  &{8.55} &{6.85}   &{11.67}  &{11.9} \\

&\multicolumn{1}{|c}{ISS}  &{6.56}   &{0.63}   &{{5.41}} &{5.53}  &{0.62}   &{0.66}  &{0.81} \\

&\multicolumn{1}{|c}{DAT}  &{12.68}  &{ 0.49}   &{{12.68}}  &{21.35}   &{0.45}  &{0.74} &{18.48} \\

&\multicolumn{1}{|c}{DivideMix}   &{31.89} &{0.47}   &{{34.04}}  &{31.77}   &{0.53}  &{0.54}  &{1.56}\\

&\multicolumn{1}{|c}{UNICON}   &\bf{42.25}&{0.93}   &{{35.48}}  &{33.37}  &{1.05}  &{1.34}  &{7.61} \\

&\multicolumn{1}{|c}{MSPL} &{{39.65}} &{\underline{{25.97}}}   &{\bf{{39.32}}}  &{\bf{{40.65}}} &{\underline{{33.33}}}  &{\underline{{30.26}}}  &{\underline{{41.87}}}  \\

&\multicolumn{1}{|c}{ERAT}  &{\underline{40.18}}  &\bf{37.23}   &\underline{{38.3}} &\underline{40.26}  &\bf{35.51} &\bf{33.85}  &\bf{42.41}   \\
\hline

\multirow{7}{*}{\tabincell{c}{\emph{Asymm.} \\\emph{$\kappa = 0.4$}}}

&\multicolumn{1}{|c}{CE}   &{0.5} &{2.55}   &{{12.35}}  &{9.76}  &{4.66}  &{7.14} &{12.42}  \\

&\multicolumn{1}{|c}{AT} &{16.92}   &{17.75}   &{{17.73}} &{15.38}  &{18.64}  &{19.76}  &{20.07}  \\

&\multicolumn{1}{|c}{ISS}  &{23.9}   &{0.53}   &{{16.88}} &{11.19}  &{0.91}   &{0.93}  &{0.58} \\

&\multicolumn{1}{|c}{DAT}  &{25.36}  &{0.63}   &{{25.36}} &{22.72}    &{0.58}  &{0.68} &{3.03} \\

&\multicolumn{1}{|c}{DivideMix}  &{36.56} &{0.56}   &{{29.94}}  &{21.3}   &{0.43}  &{0.53}  &{0.51} \\

&\multicolumn{1}{|c}{UNICON}  &{37.17} &{0.85}   &{{30.51}} &{28.09}  &{1.21}  &{1.84}  &{2.13} \\

&\multicolumn{1}{|c}{MSPL} &{\underline{{39.03}}} &{\underline{{23.78}}}   &{\underline{{38.91}}}  &{\underline{{29.38}}} &{\underline{{31.17}}}  &{\bf{{36.39}}}  &{\underline{{38.54}}}  \\

&\multicolumn{1}{|c}{ERAT}  &\bf{42.47} &\bf{26.1}   &\bf{{41.07}}  &\bf{42.96}  &\bf{33.15} &\underline{36.18}  &\bf{44.42}  \\
\toprule[1.2pt]
\end{tabular}
}
\end{table}

\subsection{\textbf{RQ1}: Robustness against dual corruption}
\label{sec:RQ1}
We use datasets corrupted by various types of label noise and data perturbations to test all methods, where the results are summarized in Table \ref{cifar10_acc}, \ref{cifar100_acc} and \ref{Tiny_acc}. In the tables, ``$\kappa$'' and ``$\epsilon^{\prime}$'' denote the magnitude of label noise and data perturbations, respectively. For label noise, the proposed method does not require any assumption on noise magnitude, while we set the defense budget $\epsilon_{\infty} = 8/255$ and $\epsilon_2= 0.5$. We also report results on clean data with clean labels in Table \ref{Tcd}. From the results, We can have the following observation. 

First, the existence of data and label corruptions poses great threats to the model robustness, degrading their performance on normal data. The compared robust learning methods tailoring for one type of corruption are not capable of dealing with the co-existence of dual corruption. In comparison, the proposed method achieves promising results and outperforms all compared methods. In particular, the proposed method can improve $8.3\%$, $13.5\%$, and $10.7\%$ robustness on CIFAR-10, CIFAR-100, and Tiny-ImageNet, respectively. These demonstrate the superiority of the proposed method. 

Second, It can be seen that different types of corruption cause different degrees of impact on the performance of the model. Specifically, in terms of label corruption, instance-level label noise poses a more significant threat to model robustness. On CIFAR-10, the compared methods achieved an average best performance of $76.1 (\%)$ in instance-dependent label noise, which is below $79.3 (\%)$ and $81.36 (\%)$ achieved in symmetric and asymmetric label noise. On CIFAR-100, the accuracy is $47.02(\%), 54.15(\%)$ and $44.37 (\%)$, respectively. On Tiny-ImageNet, the average best performance of the compared methods is $31.17 (\%), 36.09 (\%)$ and $33.89 (\%)$. The main reason may be that the instance-dependent label noise is generated based on both labels and features, thus being more delusive. By contrast, the proposed method has an average performance of $84.24 (\%)$ and $53.21 (\%)$ when facing instance-dependent label noise on CIFAR-10 and CIFAR-100, which are similar to the performance in symmetric label noise, i.e., $84.65$ (\%) and $52.99 (\%)$, and asymmetric label noise, i.e., $86.01$ (\%) and $58.46 (\%)$. An exception is the performance on Tiny-ImageNet, where the proposed ERAT achieved an average performance of $32.98 (\%)$ in instance-dependent label noise which is inferior to $49.41 (\%)$ in symmetric label noise and $38.05 (\%)$ in asymmetric label noise. Overall, the proposed ERAT can effectively defend against label corruption.  

Lastly, it is noted that in some cases, e.g., \emph{UAP + Inst.}, the MSPL achieves slightly better performance than the proposed method.  A potential reason is that it assumes the corruption types are all available. However, the proposed method can achieve the overall best results of defending against different types of data and label noise without knowing noise information. This further demonstrates its practicability, which is very important for real-world applications. 

\begin{figure*}[tbh]
\begin{subfigure}[\emph{L2C + Inst.}]
{\includegraphics[angle=0, width=0.23\textwidth]{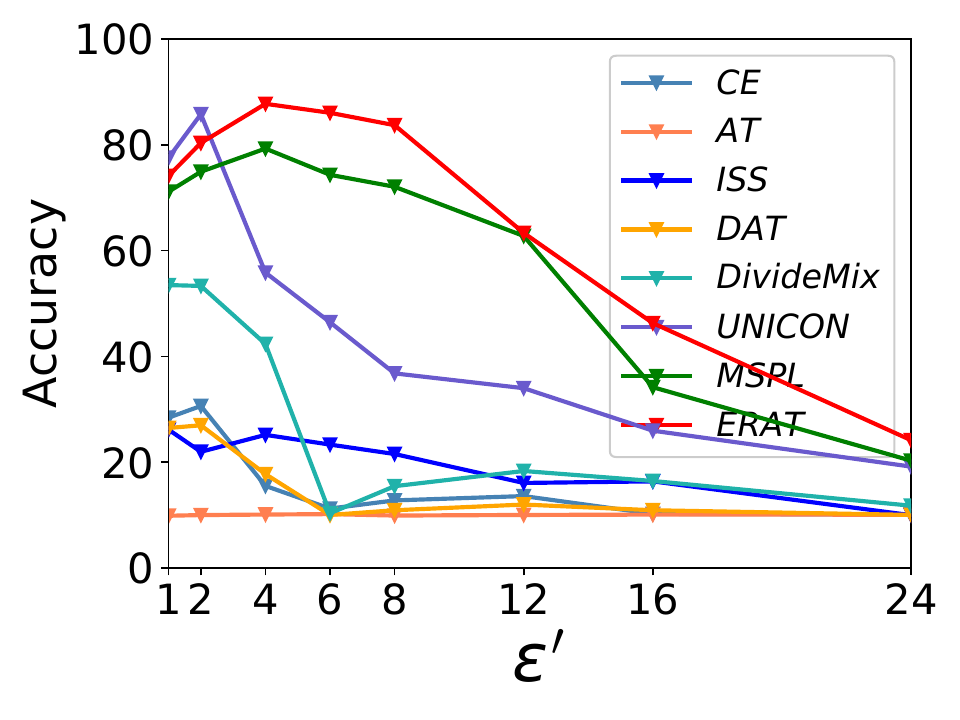}}
\end{subfigure}
\begin{subfigure}[\emph{L2C + Symm.}]
{\includegraphics[angle=0, width=0.23\textwidth]{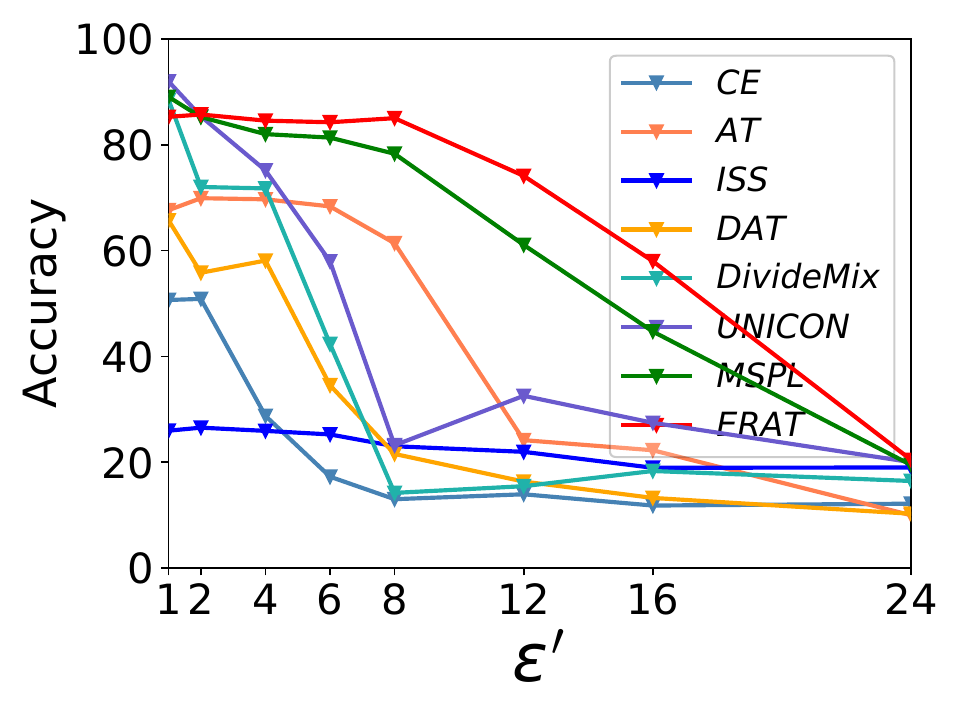}}
\end{subfigure}
\begin{subfigure}[\emph{UAP + Inst.}]
{\includegraphics[angle=0, width=0.23\textwidth]{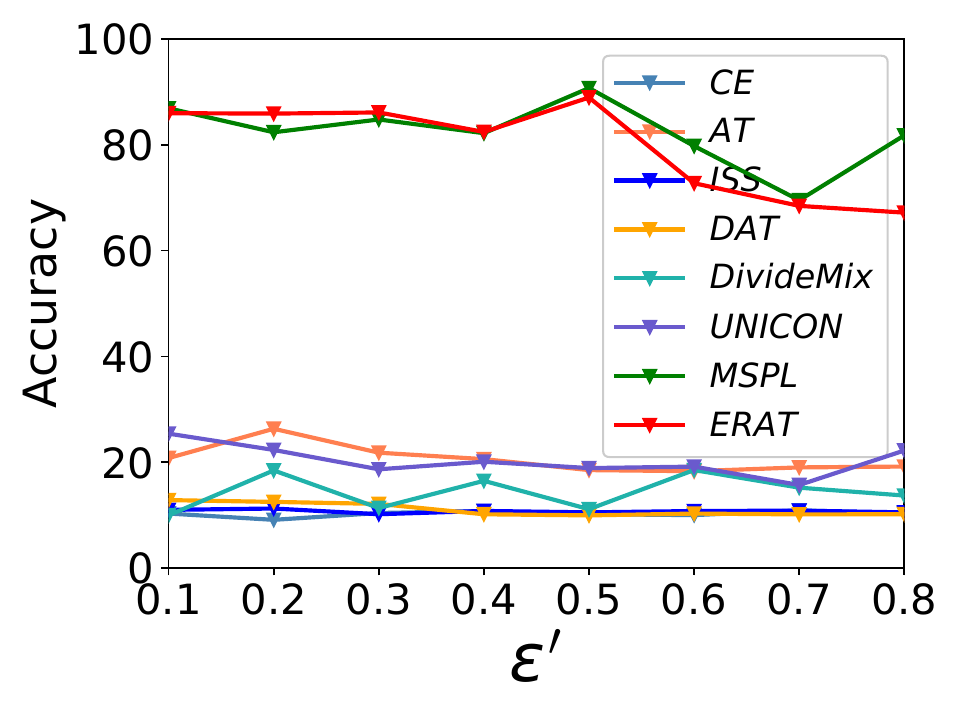}}
\end{subfigure}
\begin{subfigure}[\emph{UAP + Symm.}]
{\includegraphics[angle=0, width=0.23\textwidth]{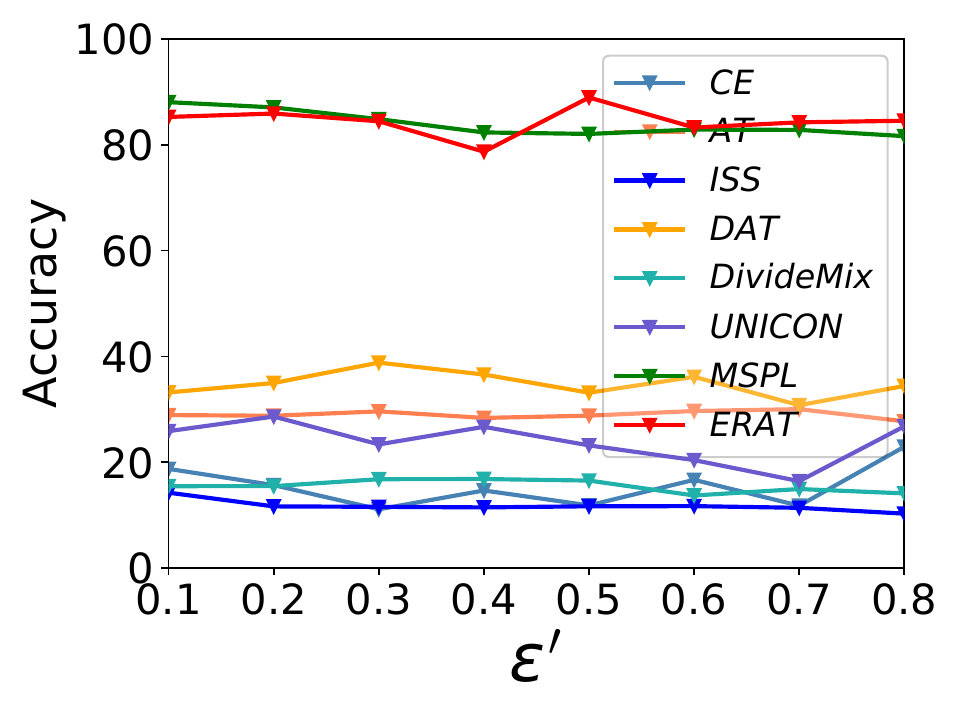}}
\end{subfigure}\\
\caption{Test accuracy on CIFAR-10 under different magnitudes of data corruption. When $\epsilon^{\prime}$ approaches 0, it means that there is only label noise. In this case, the proposed method still performs better than methods that are primarily designed for defending against label noise, demonstrating the applicability of the proposed method.}
\label{datamag_acc}
\end{figure*}

\begin{figure*}[tbh]
\begin{subfigure}[\emph{L2C + Inst.}]
{\includegraphics[angle=0, width=0.23\textwidth]{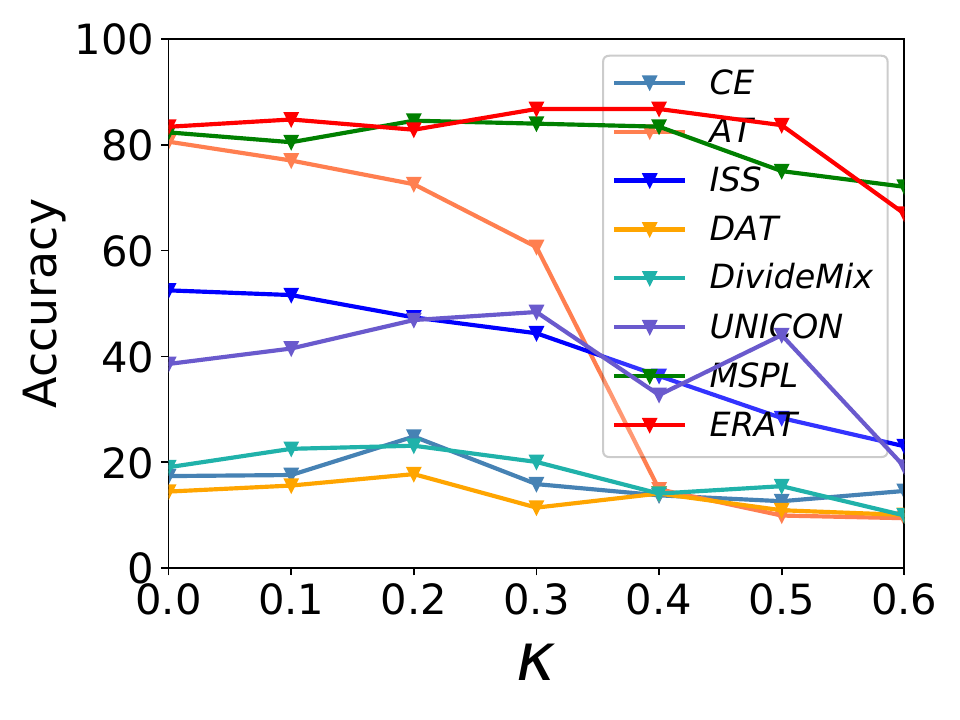}}
\end{subfigure}
\begin{subfigure}[\emph{L2C + Symm.}]
{\includegraphics[angle=0, width=0.23\textwidth]{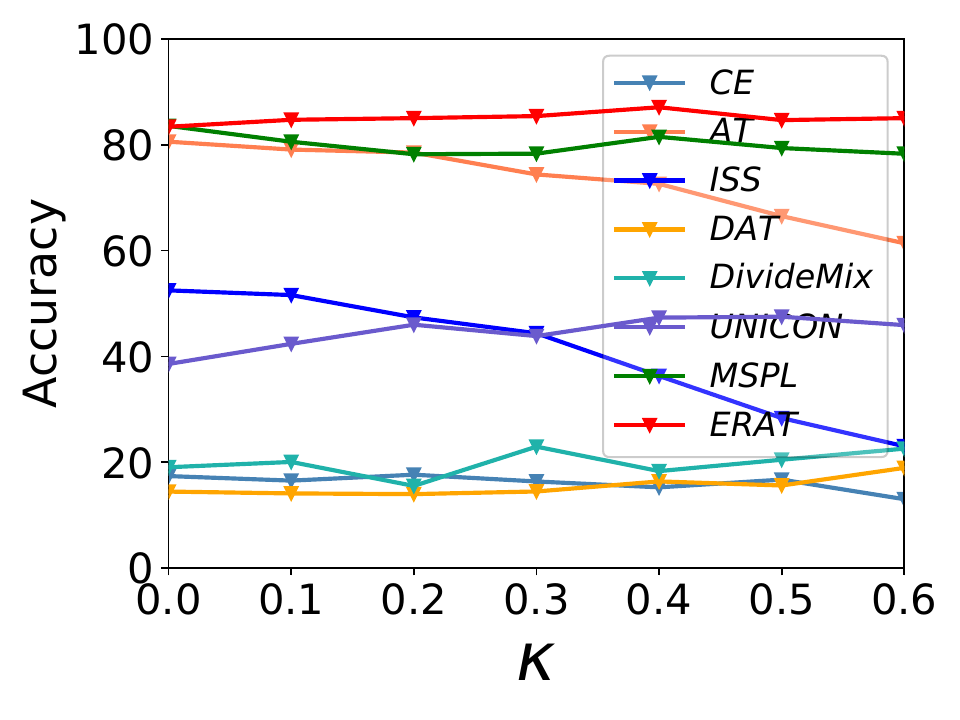}}
\end{subfigure}
\begin{subfigure}[\emph{UAP + Inst.}]
{\includegraphics[angle=0, width=0.23\textwidth]{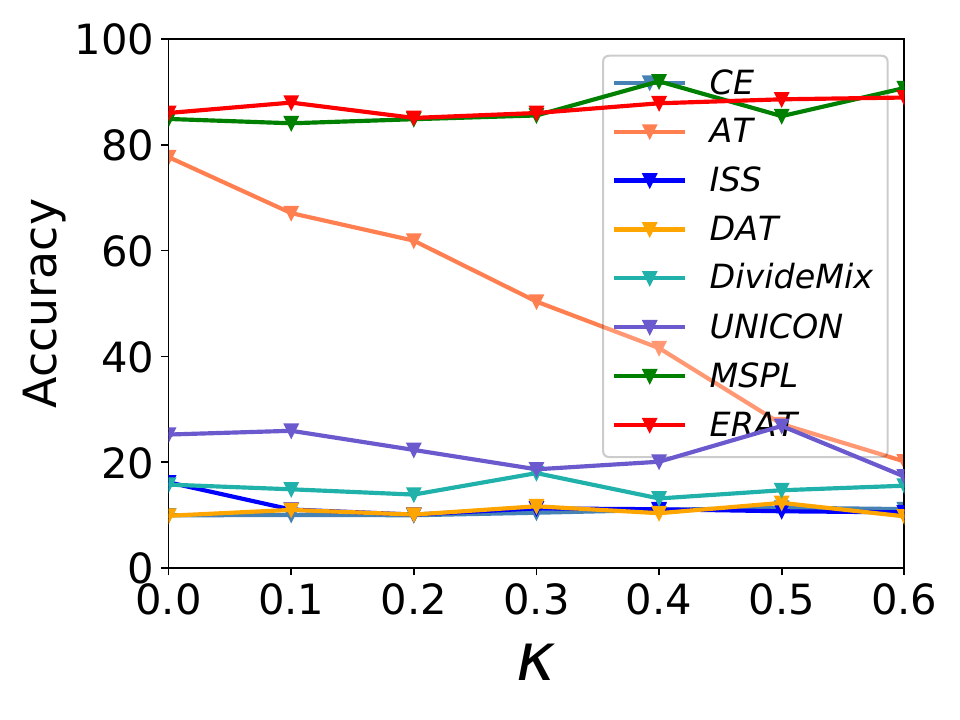}}
\end{subfigure}
\begin{subfigure}[\emph{UAP + Symm.}]
{\includegraphics[angle=0, width=0.23\textwidth]{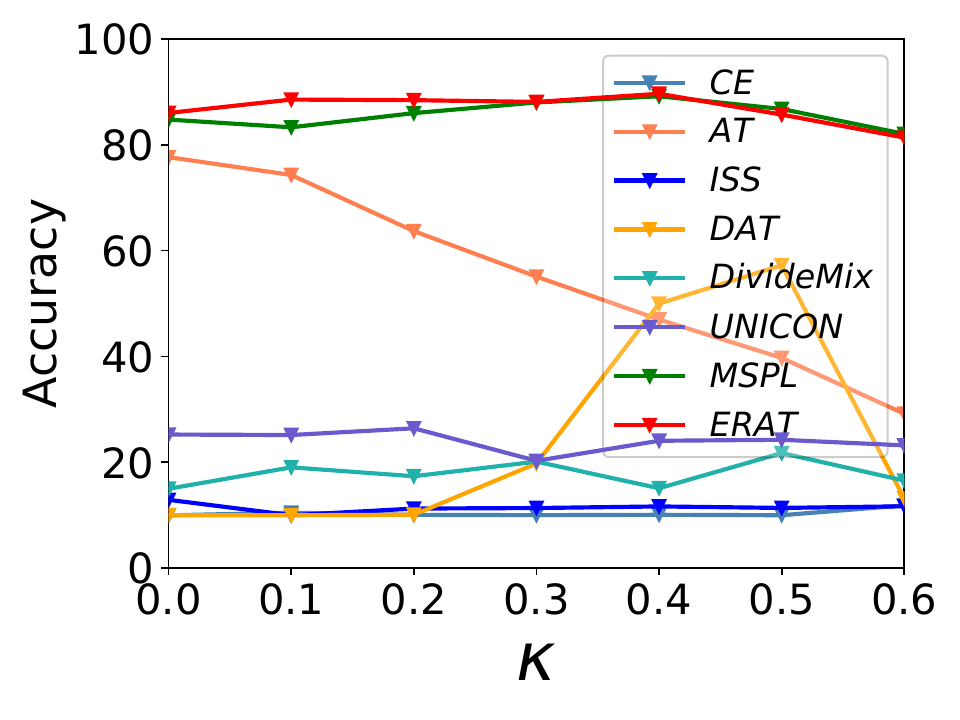}}
\end{subfigure}\\
\caption{Test accuracy on CIFAR-10 under different magnitudes of label corruption. When $\kappa$ approaches 0, it means that there are only data perturbations. In this case, the proposed method still performs better than methods that are primarily designed for defending data noise, demonstrating the applicability of the proposed method.}
\label{labelmag_acc}
\end{figure*}

\subsection{\textbf{RQ2}: Varying the corruption magnitude and defense budget}
Here, we study the model performance under different corruption magnitudes. The results are presented in Figure \ref{datamag_acc} and \ref{labelmag_acc}, revealing the following observations.

In terms of data corruption, it can be seen that the performance of all compared methods would be influenced significantly by the existence of data perturbations, though the influence varies from each other. Concretely, as the noise magnitude increases, $l_{\infty}$-norm data perturbations would cause more serious threats to models while small $l_{2}$-norm data perturbations would lead to the inefficacy of models. Promisingly, the proposed method achieves the best performance and significantly improves the robustness of the model. This further verifies the effectiveness of the proposed method. For label corruption, similar results can be observed. Larger label noise would cause worse performance of models. However, our method can have a stable and prominent performance. 

It is worth noting that a small corruption magnitude, e.g., $\epsilon' \rightarrow 2/255$ or $\epsilon' \rightarrow 0.1$ refers to almost labels noise only or data perturbations only respectively. These scenarios are particularly designed to make a fair performance study between our proposed method and the baseline methods that are only developed for one type of corruptions.  

The experimental results evidence the robustness of the ERAT in both extreme cases, which means that it is constantly effective in defending against the co-existence of corruption or just one type of them, further verifying its applicability. 

\begin{figure*}[tbh]
\begin{subfigure}[\emph{L2C + Inst.}]
{\includegraphics[angle=0, width=0.23\textwidth]{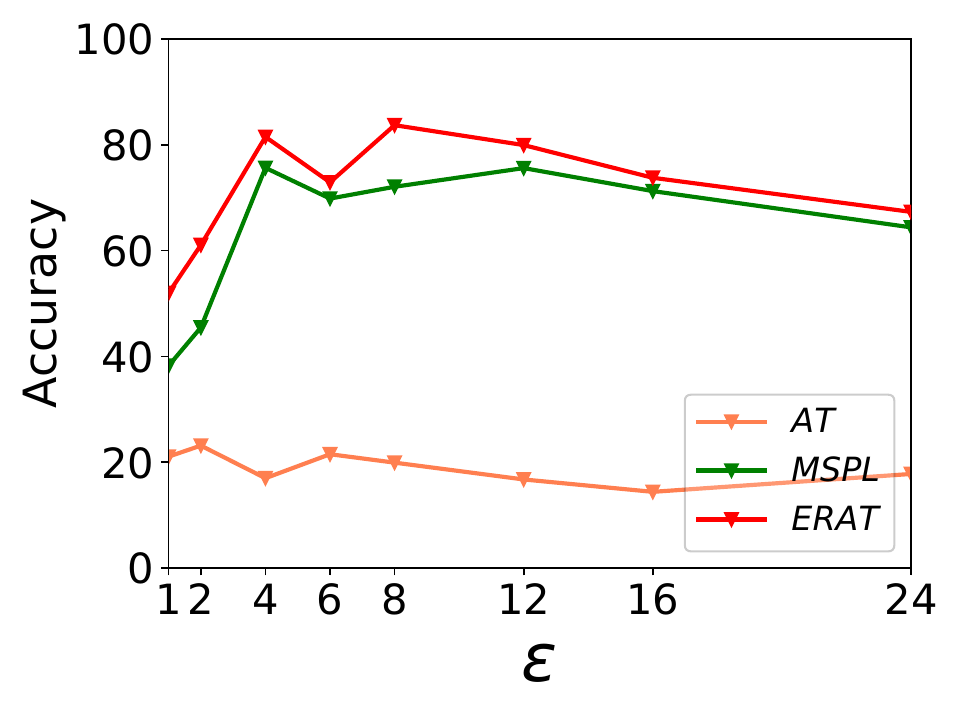}}
\end{subfigure}
\begin{subfigure}[\emph{L2C + Symm.}]
{\includegraphics[angle=0, width=0.23\textwidth]{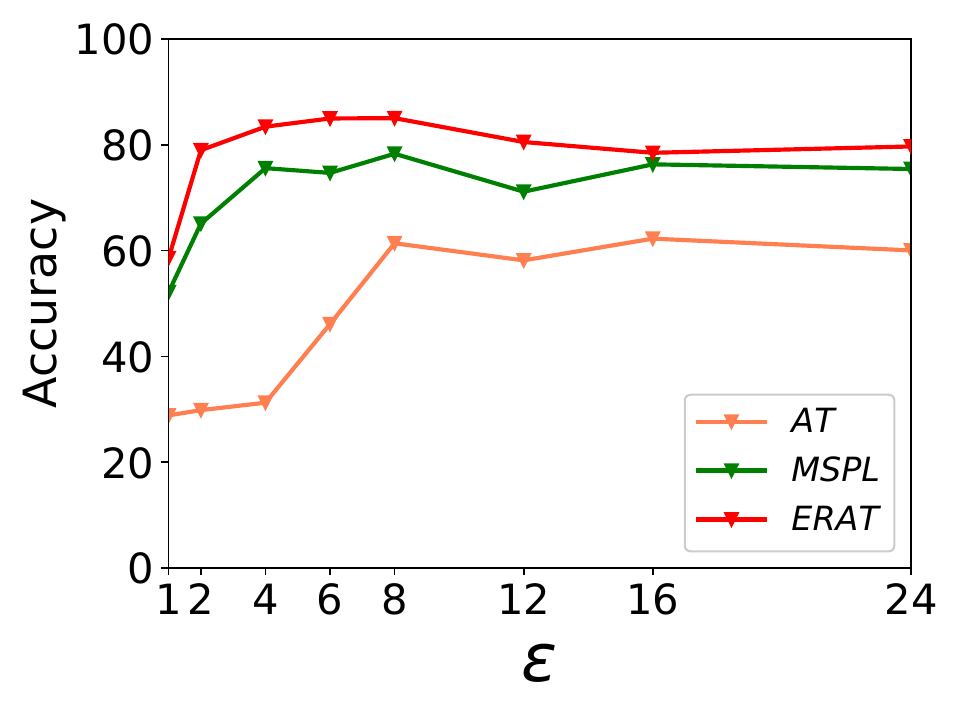}}
\end{subfigure}
\begin{subfigure}[\emph{UAP + Inst.}]
{\includegraphics[angle=0, width=0.23\textwidth]{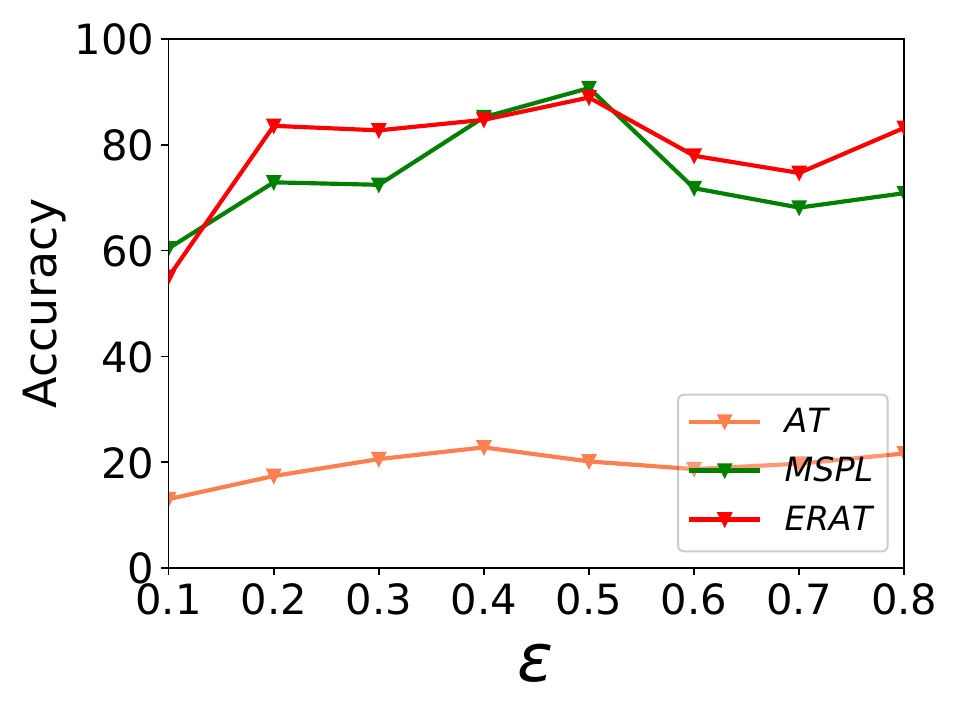}}
\end{subfigure}
\begin{subfigure}[\emph{UAP + Symm.}]
{\includegraphics[angle=0, width=0.23\textwidth]{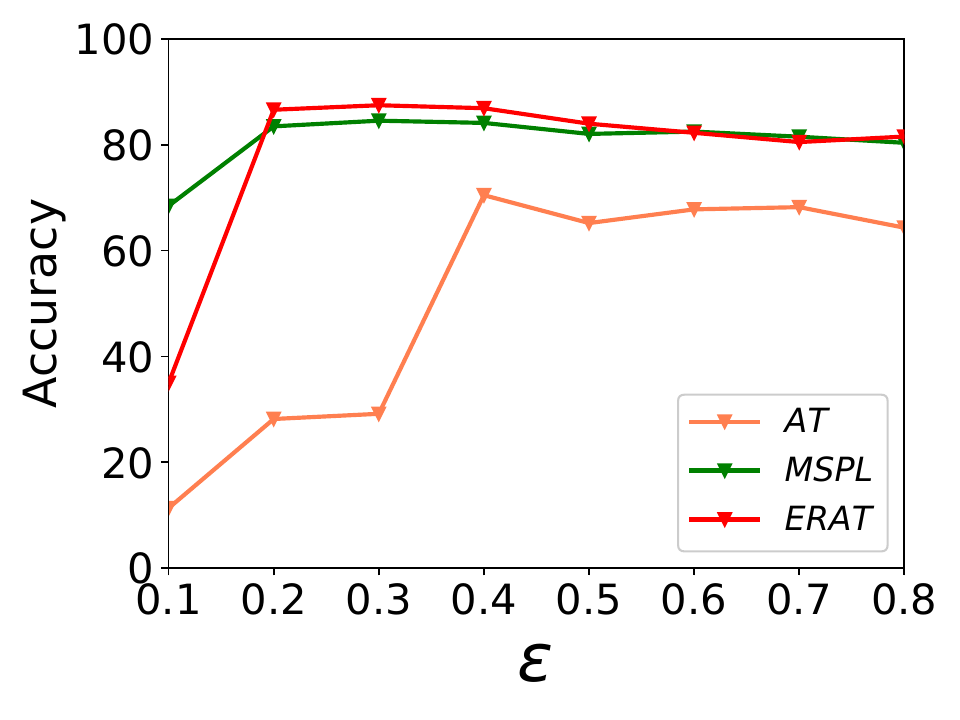}}
\end{subfigure}\\
\caption{Test accuracy on CIFAR-10 with different defense budgets against data corruption.}
\label{bg_acc}
\end{figure*}

More experiments are conducted to further investigate the influence of the defense budget, where the results are reported in Figure \ref{bg_acc}. In the figure, ``1, 2, 4, 6, 8, 16, 24'' represent the budget of $(\epsilon_{\infty}, \epsilon_{2}) =(1/255,0.1), (2/255,0.2)$, $(4/255,0.3)$, $(6/255,0.4)$, $(8/255,0.5)$, $(12/255,0.6)$, $(16/255,0.7)$, $(24/255,0.8)$. It can be seen that a too-large or a too-small budget would influence the defense effects, which indicates that setting a proper budget is vitally important for effective defense. At the same time, the threat of instance-dependent label noise is further verified. The compared AT has better performance in symmetric label noise than instance-dependent label noise.

\begin{figure*}[tbh]
\begin{subfigure}[\emph{L2C + Inst.}]
{\includegraphics[angle=0, width=0.23\textwidth]{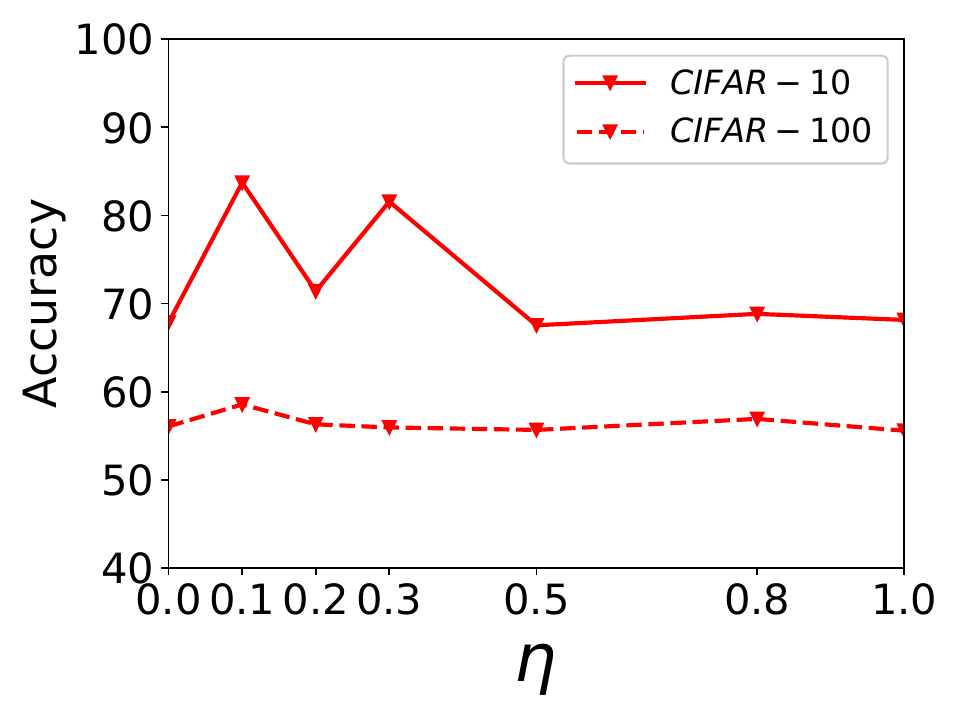}}
\end{subfigure}
\begin{subfigure}[\emph{L2C + Symm.}]
{\includegraphics[angle=0, width=0.23\textwidth]{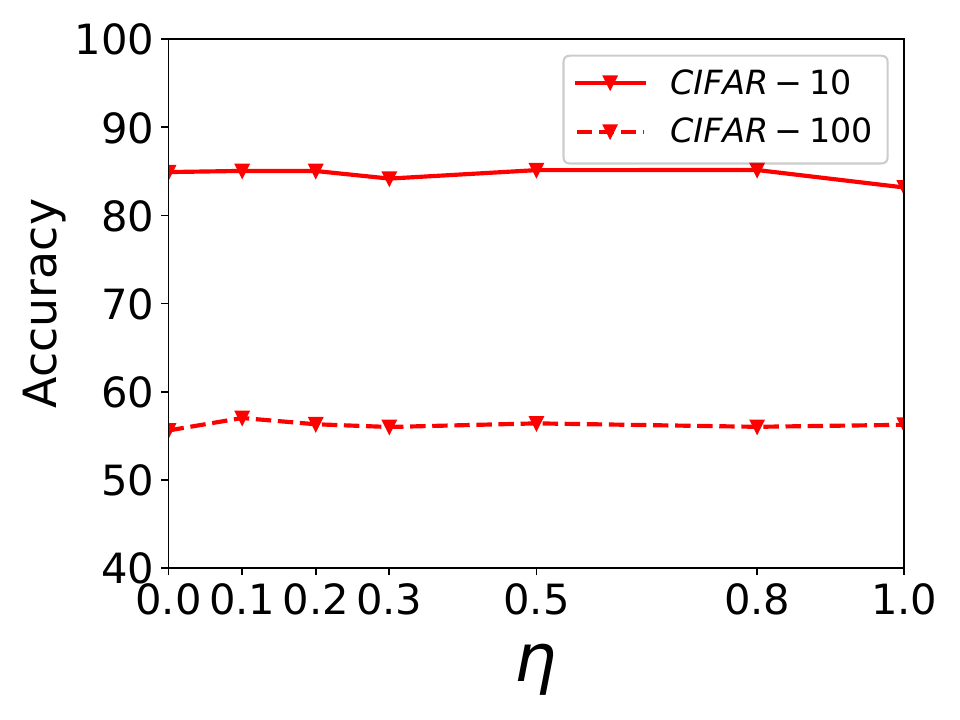}}
\end{subfigure}
\begin{subfigure}[\emph{UAP + Inst.}]
{\includegraphics[angle=0, width=0.23\textwidth]{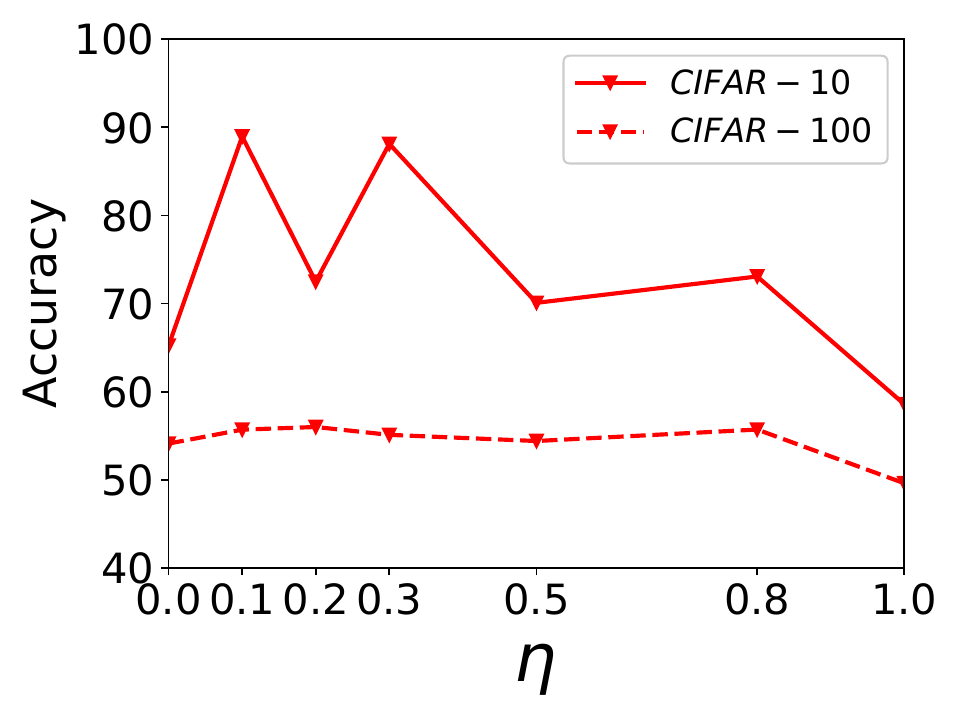}}
\end{subfigure}
\begin{subfigure}[\emph{UAP + Symm.}]
{\includegraphics[angle=0, width=0.23\textwidth]{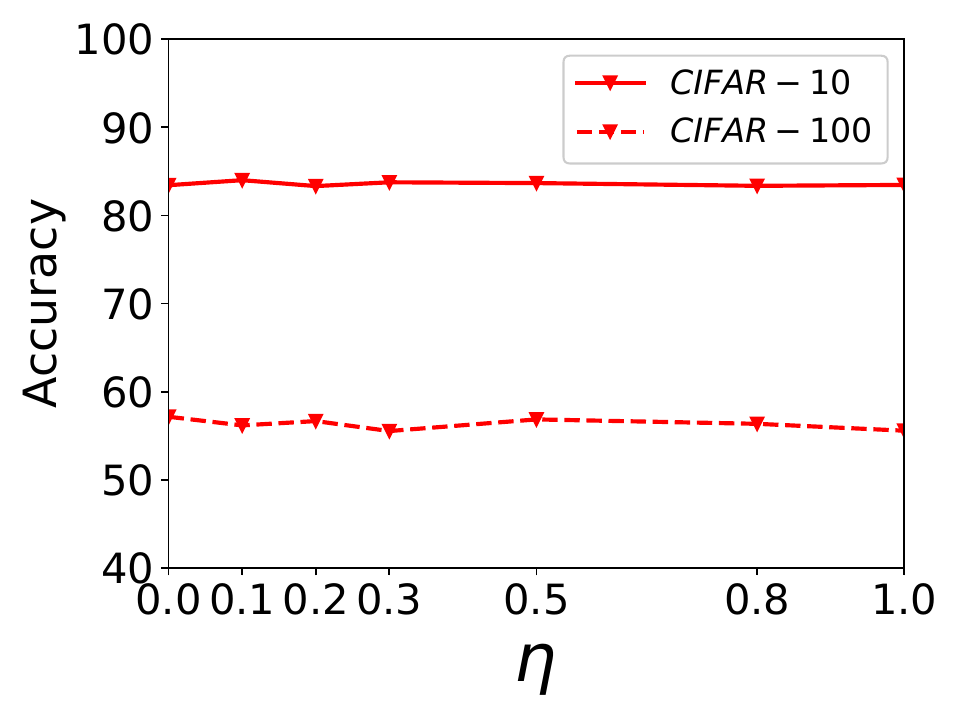}}
\end{subfigure}\\
\caption{Test accuracy with different rebalancing strength on CIFAR-10.}
\label{eta_acc}
\end{figure*}

\begin{figure*}[tbh]
\begin{subfigure}[\emph{L2C + Inst.}]
{\includegraphics[angle=0, width=0.23\textwidth]{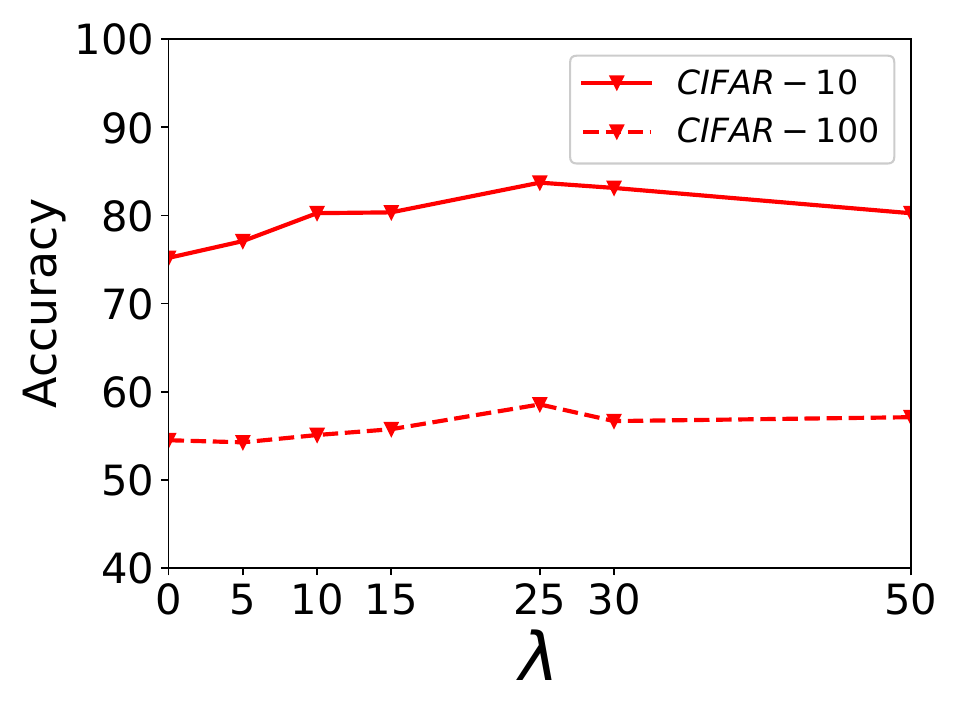}}
\end{subfigure}
\begin{subfigure}[\emph{L2C + Symm.}]
{\includegraphics[angle=0, width=0.23\textwidth]{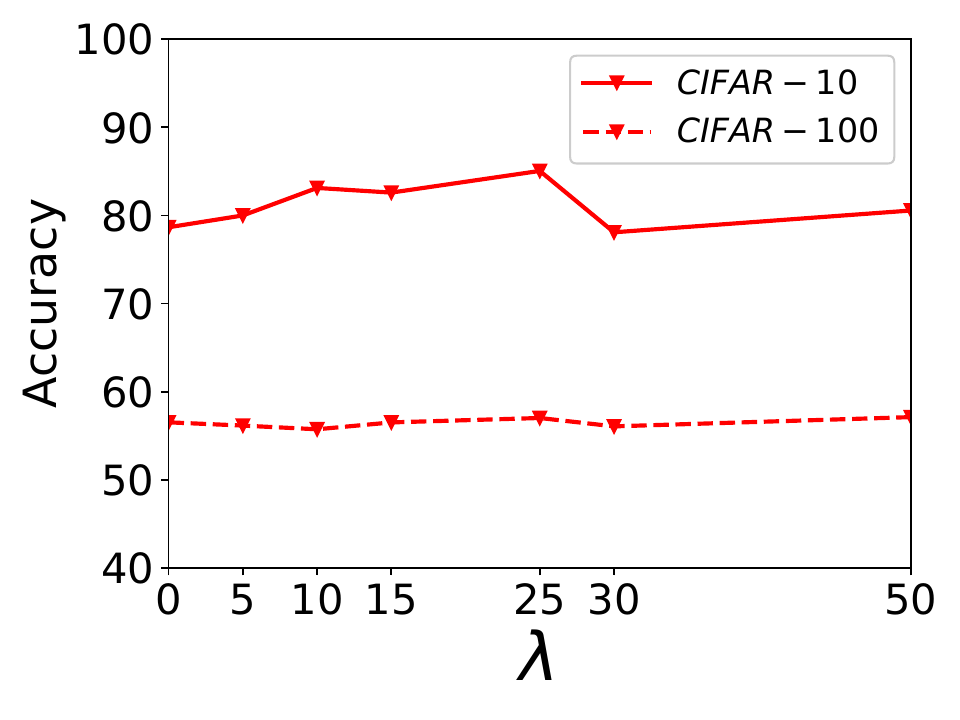}}
\end{subfigure}
\begin{subfigure}[\emph{UAP + Inst.}]
{\includegraphics[angle=0, width=0.23\textwidth]{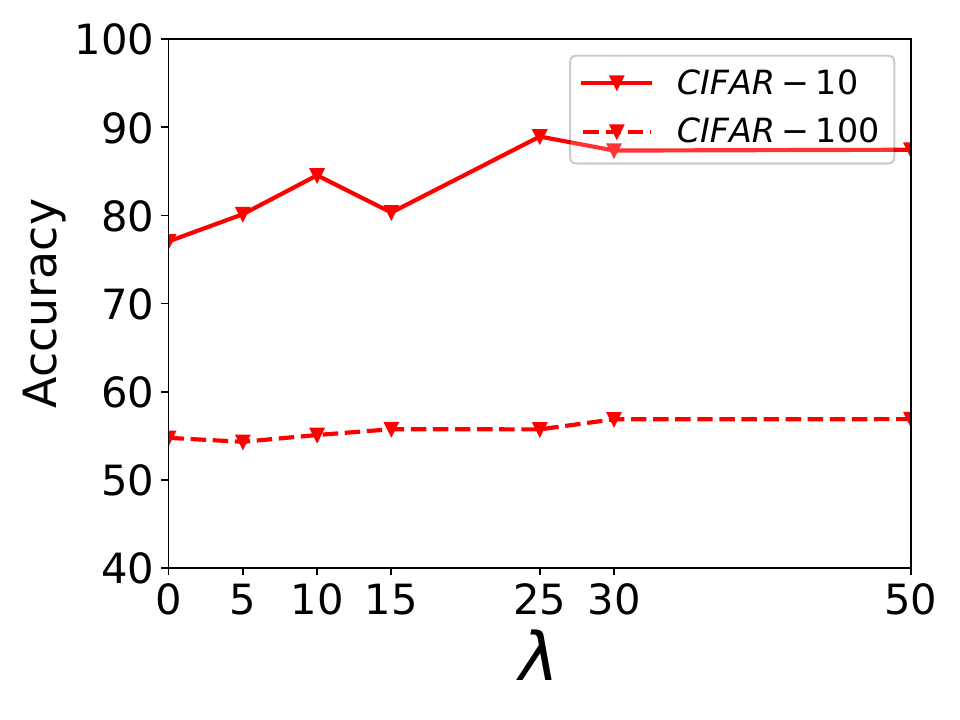}}
\end{subfigure}
\begin{subfigure}[\emph{UAP + Symm.}]
{\includegraphics[angle=0, width=0.23\textwidth]{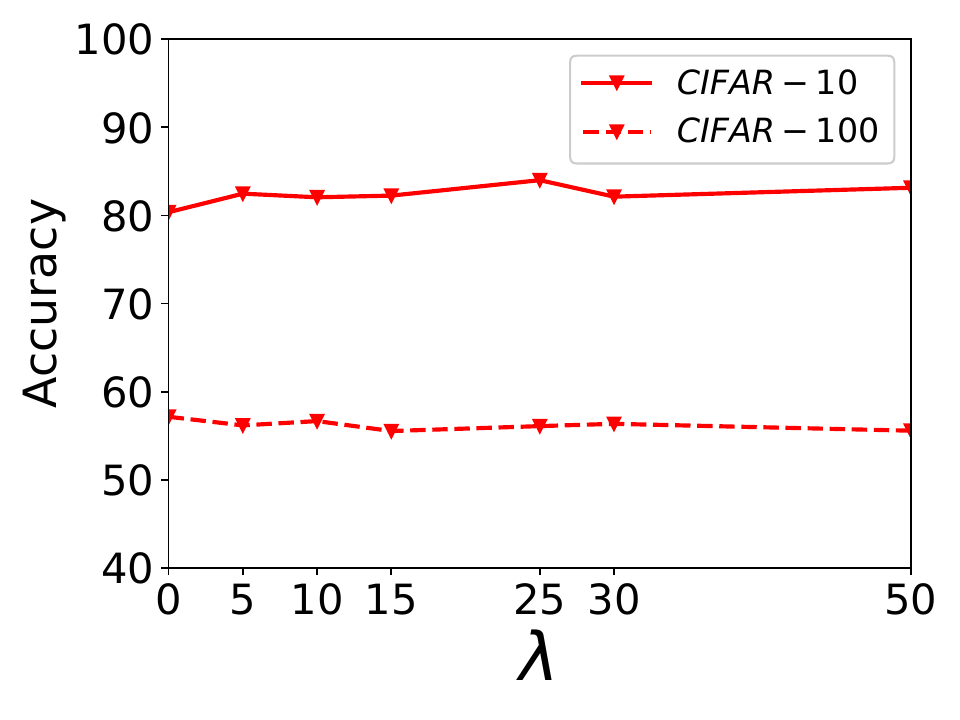}}
\end{subfigure}\\
\caption{Analysis of parameter $\lambda$ on CIFAR-10.}\medskip
\label{lam_acc}
\end{figure*}

\begin{figure*}[tbh] 
\begin{subfigure}[\emph{L2C + Inst.}]
{
\includegraphics[angle=0, width=0.16\textwidth]{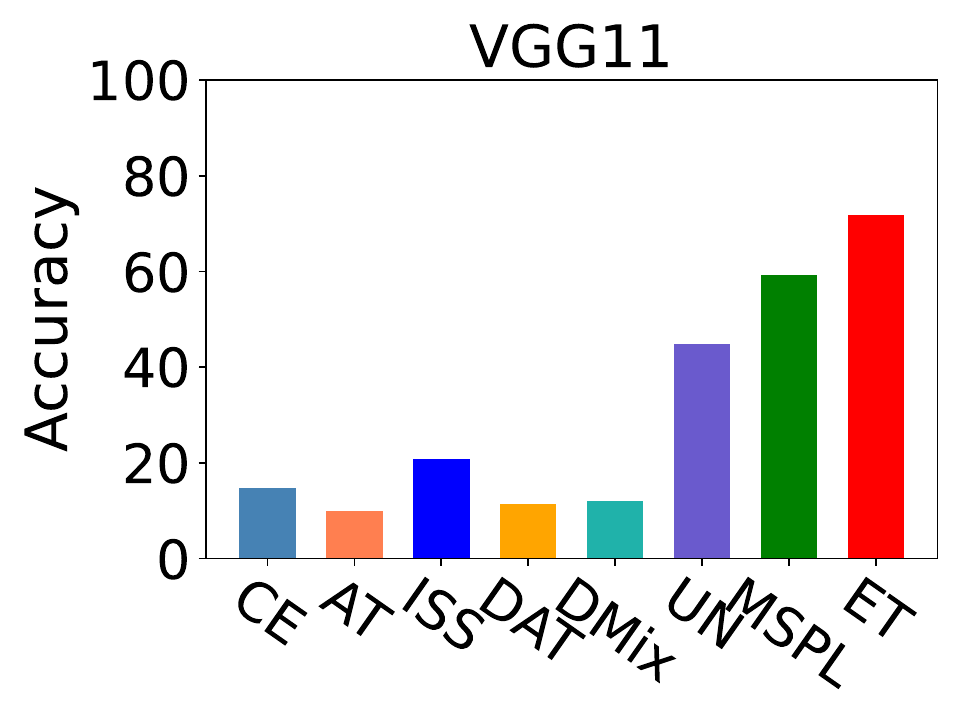}
\includegraphics[angle=0, width=0.16\textwidth]{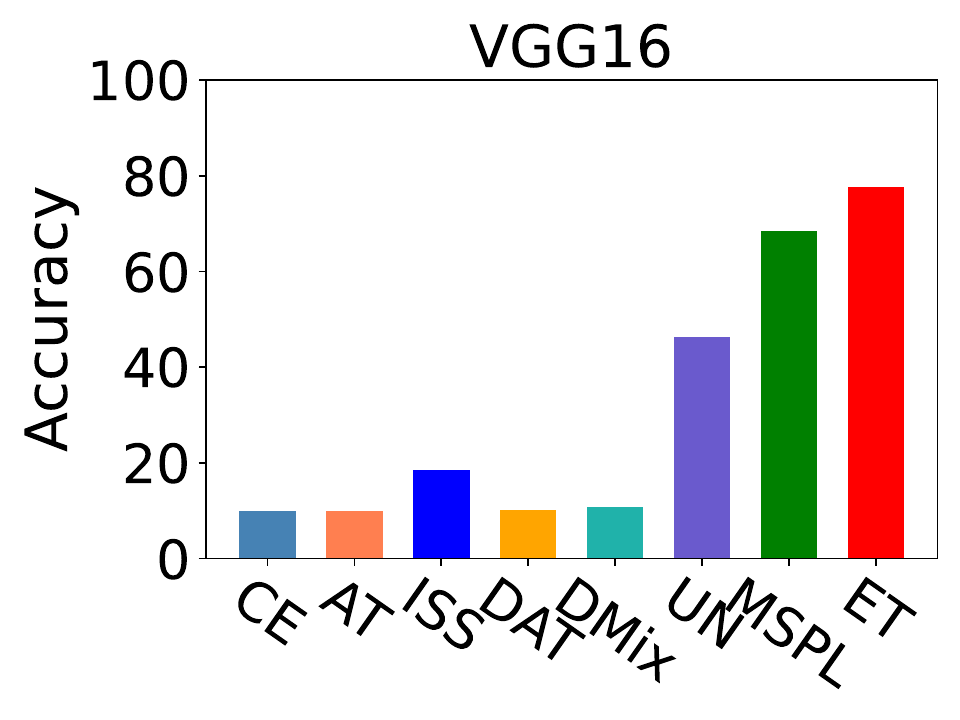}
\includegraphics[angle=0, width=0.16\textwidth]{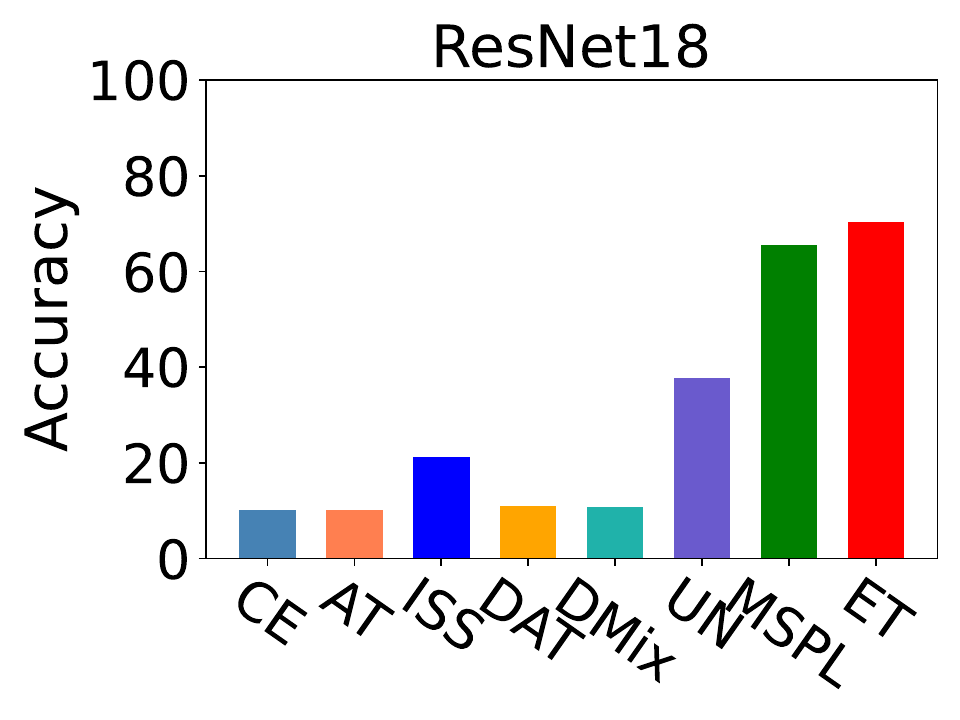}\\
\includegraphics[angle=0, width=0.16\textwidth]{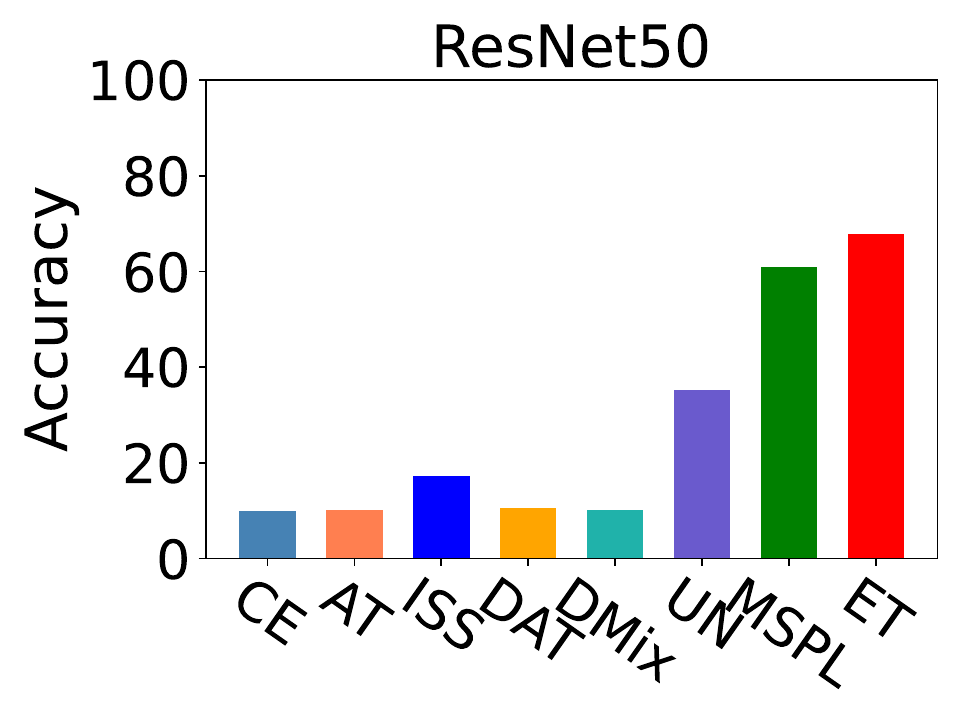}
\includegraphics[angle=0, width=0.16\textwidth]{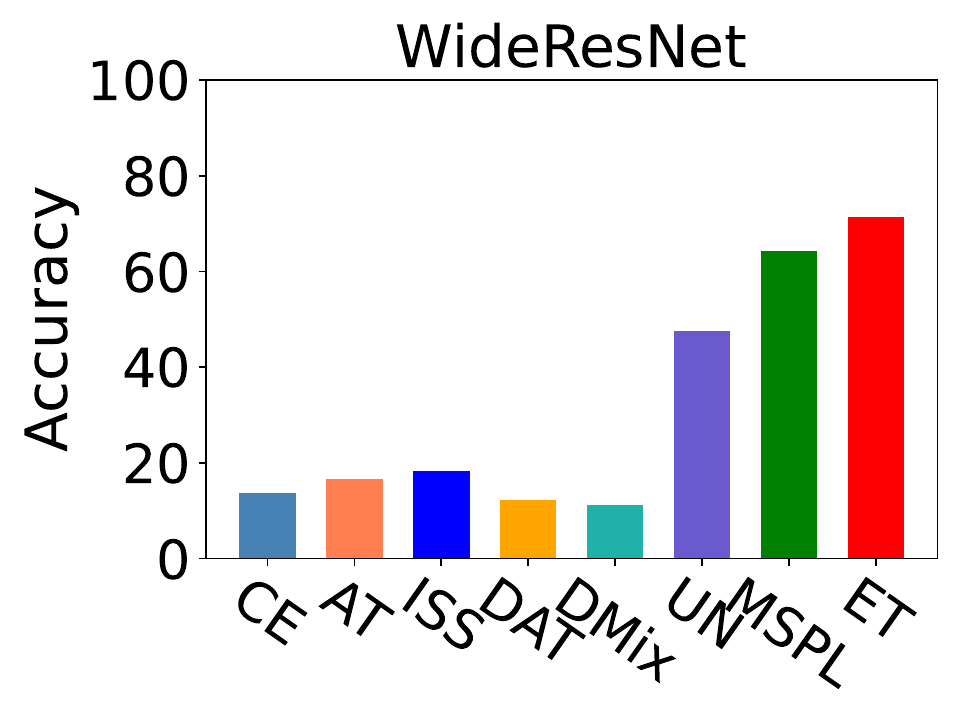}
\includegraphics[angle=0, width=0.16\textwidth]{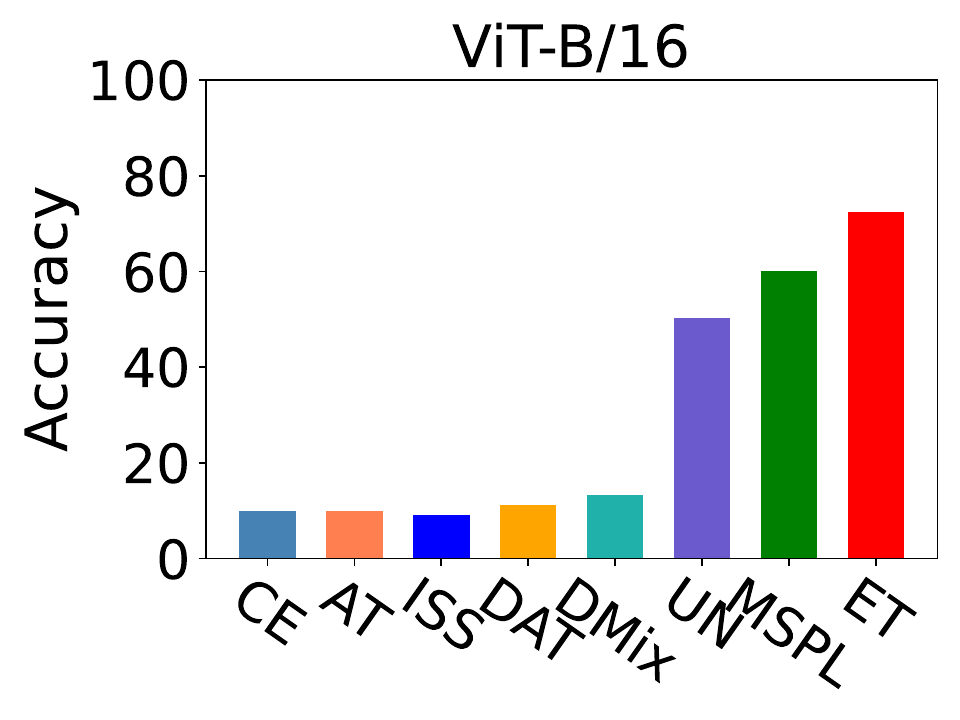}}
\end{subfigure}
\begin{subfigure}[\emph{L2C + Symm.}]
{
\includegraphics[angle=0, width=0.16\textwidth]{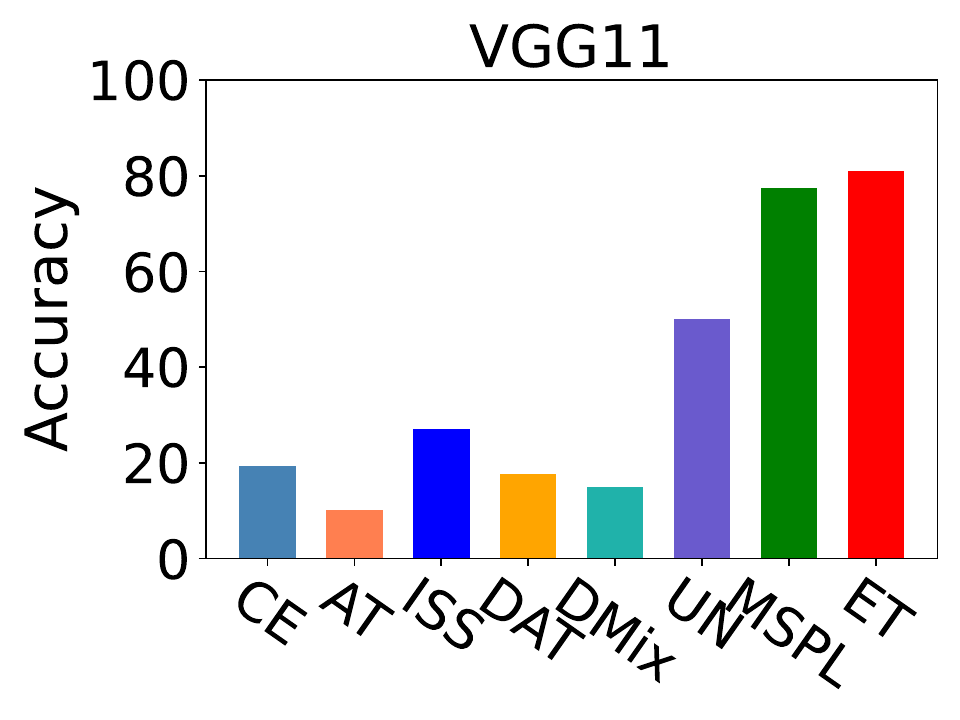}
\includegraphics[angle=0, width=0.16\textwidth]{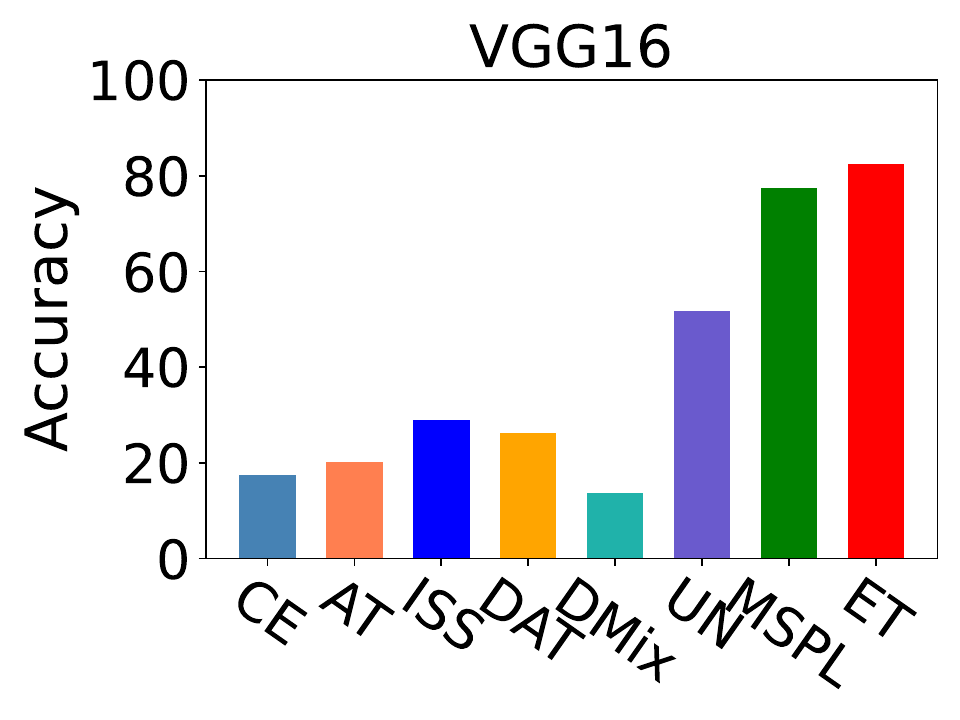}
\includegraphics[angle=0, width=0.16\textwidth]{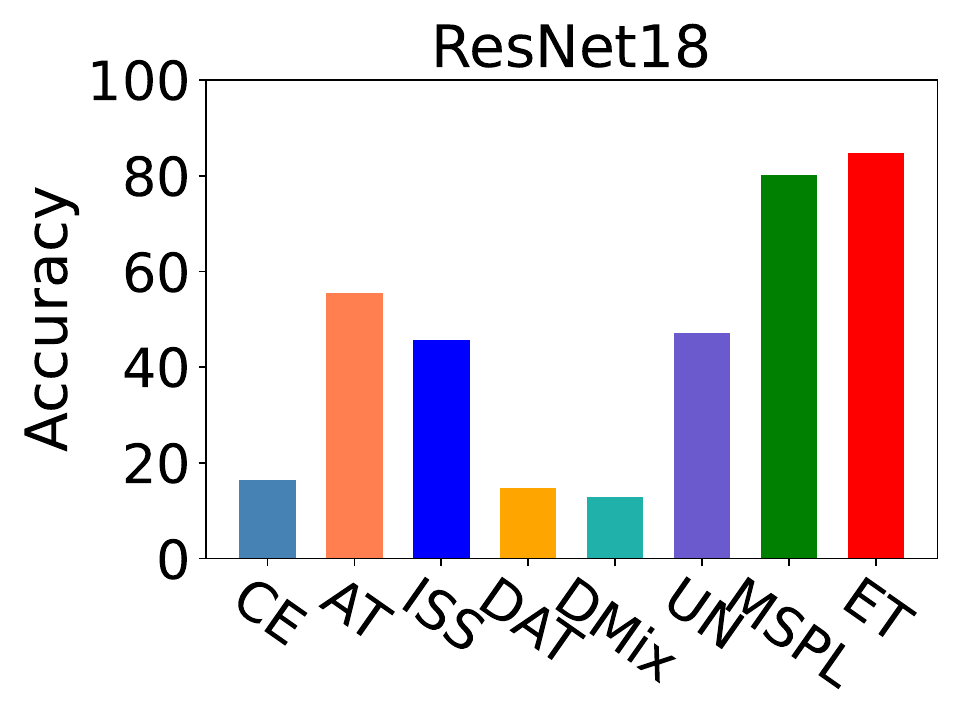}\\
\includegraphics[angle=0, width=0.16\textwidth]{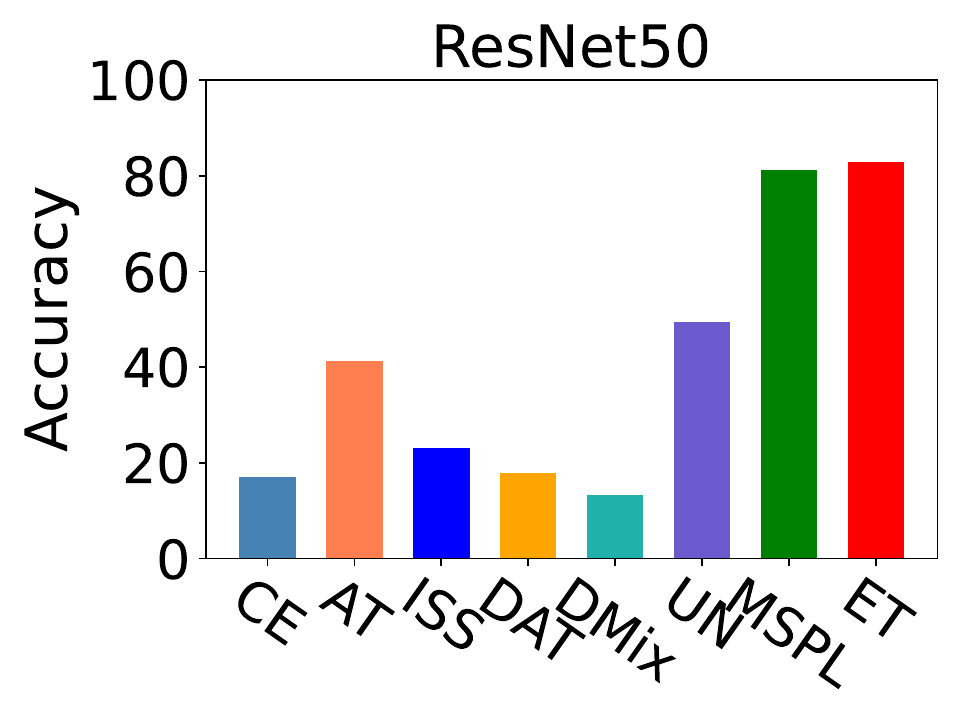}
\includegraphics[angle=0, width=0.16\textwidth]{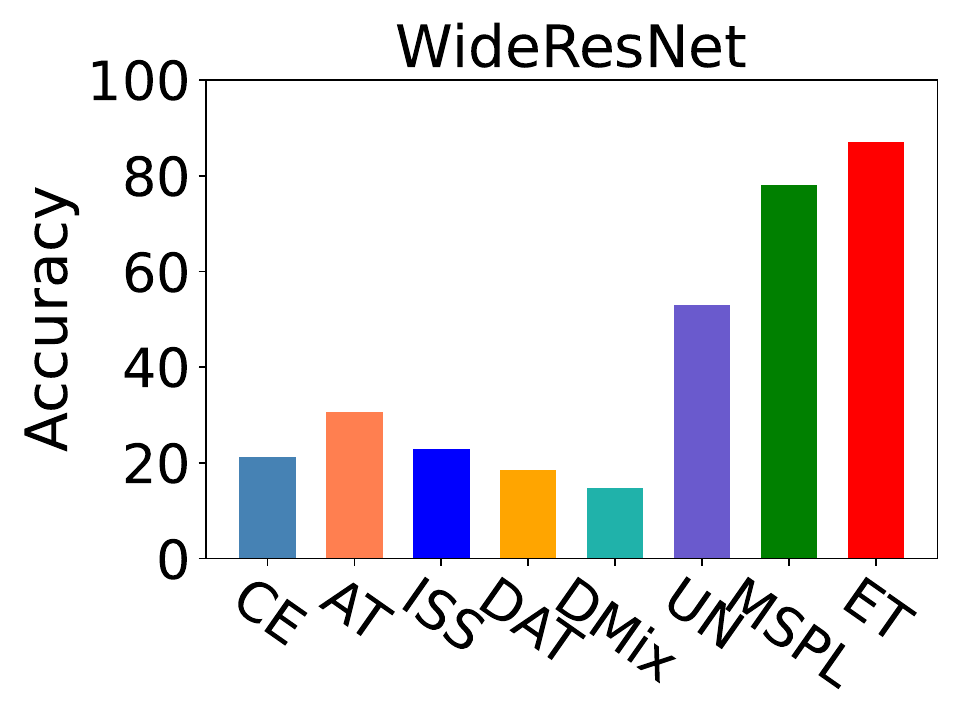}
\includegraphics[angle=0, width=0.16\textwidth]{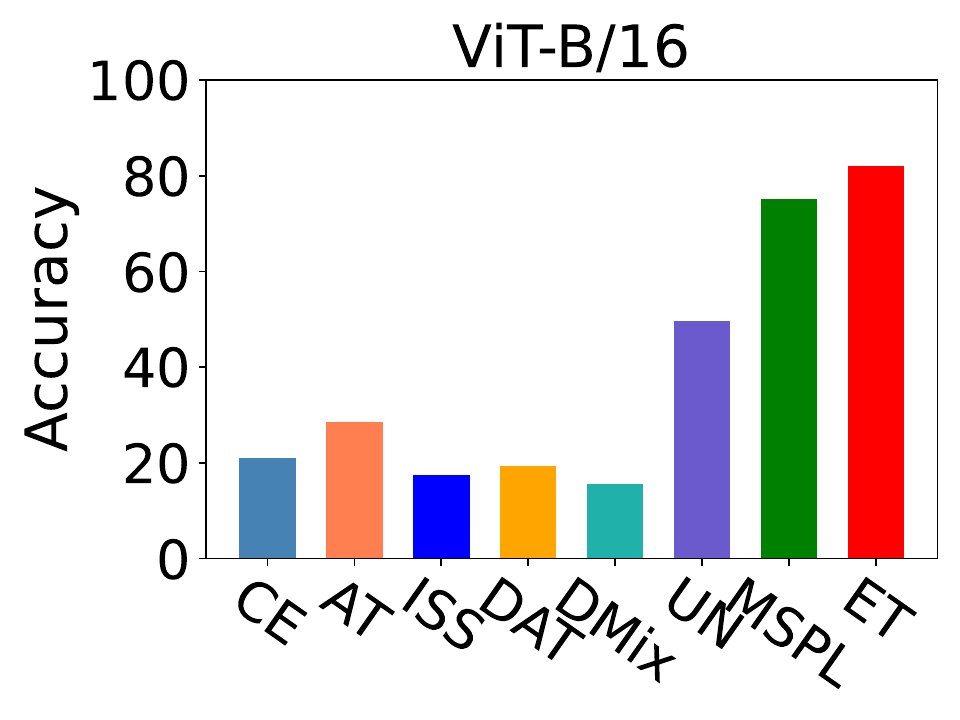}}
\end{subfigure}
\caption{Test accuracy with different networks on CIFAR-10. For better representation, we abbreviate methods ``DivideMix'', ``UNICON'', ``ERAT'' with ``DMix'', ``UN'', ``ET''. }
\label{bb_acc}
\end{figure*}

\subsection{\textbf{RQ3}: Effectiveness on different networks}
In real-world applications, classification models are often built on different networks. So, it is vitally important to test whether the proposed method can work on different networks. To this end, we select various networks including VGG11, VGG16, ResNet18, ResNet50, WideResNet, and ViT-B$/$16 for all learning methods. Results on CIFAR-10 are reported in Figure \ref{bb_acc}. From the figures, it can be observed the threats caused by data and label noise to all compared methods. In the meanwhile, it is shown that our defense method is effective on different architectures.

\subsection{\textbf{RQ4}: Importance of each component}

\begin{table}[tb]
\center
\caption{Ablation study results in terms of test accuracy $(\%)$ on CIFAR-10. \textbf{Bold} indicates the best and the second best is \underline{underlined}.}\label{alba}
\scalebox{0.75}{
\begin{tabular}{ccc cccc cccc cccc cccc}\toprule[1pt]
\multirow{2}{*}{\emph{Corruption}} &\multirow{2}{*}{{Method}} &\multirow{1}{*}
{\emph{L2C}} &\multirow{1}{*}{\emph{UP}} &\multirow{1}{*}{\emph{AA}} &\multirow{1}{*}{\emph{AP}}&\multirow{1}{*}{\emph{UAP}} &\multirow{1}{*}{\emph{UHP}}&\multirow{1}{*}{\emph{URP}} \\

& &\multicolumn{2}{c}{\emph{$l_{\infty},\epsilon^{\prime} =8/255$}} &\multicolumn{4}{c}{\emph{$l_{2},\epsilon^{\prime} =0.5$}}\\ \toprule[1pt] 

\multirow{6}{*}{\tabincell{c}{\emph{Inst}. \\\emph{$\kappa = 0.6$}}}
&\multicolumn{1}{|c}{ERAT}  &{\underline{83.71}}   &{\underline{69.93}} &{\underline{85.66}} &{86.1} &{\textbf{88.94}}  &{\textbf{86.16}}   &{\textbf{89.15}} \\

&\multicolumn{1}{|c}{ERAT$_{\infty}$}    &{\textbf{84.35}}   &{70.37}  &{81.01} &{\underline{88.33}} &{28.16}  &{18.18}   &{77.34} \\

&\multicolumn{1}{|c}{ERAT$_2$}     &{41.12}   &{13.27} &{60.36} &{85.07} &{\underline{84.43}}  &{\underline{78.97}}   &{85.69} \\

&\multicolumn{1}{|c}{ERAT w/ CE}  &{73.88}   &{61.21}  &{75.57} &{78.19}&{68.67}  &{70.07}   &{79.65} \\

&\multicolumn{1}{|c}{ERAT w/o RB}  &{10.01}   &{10.13}  &{10.06} &{9.87} &{10.2}  &{12.2}   &{11.17} \\ 

&\multicolumn{1}{|c}{ERAT w/o Uni}  &{83.03}   &{\textbf{75.48}} &\bf{86.72}  &{\textbf{88.82}} &{70.34}  &{69.24}   &{\underline{86.57}} \\

&\multicolumn{1}{|c}{ERAT w/o Aug} &{{73.04}} &{{63.65}}   &{{70.37}}  &{72.47} &{{72.28}} &{{75.33}} &{{75.85}}  \\

&\multicolumn{1}{|c}{ERAT w/o AT} &{{31.99}} &{{11.12}}  &{{48.24}} &{{66.43}} &{{10.46}} &{{11.46}} &{{85.28}}  \\

&\multicolumn{1}{|c}{ERAT w/o SS} &{{17.72}} &{{14.61}}  &{{22.28}} &{{17.08}} &{{12.11}} &{{22.36}} &{{17.64}}  \\\hline

\multirow{6}{*}{\tabincell{c}{\emph{Symm}. \\\emph{$\kappa = 0.6$}}}
&\multicolumn{1}{|c}{ERAT}  &{\textbf{85.04}} &{\textbf{82.31}} &\bf{85.51} &{85.84} &{83.99}  &{83.44}  &{86.44}  \\

&\multicolumn{1}{|c}{ERAT$_{\infty}$}    &{80.97}   &{70.46} &{82.82} &{\textbf{88.18}} &{57.52}  &{50.75} &{\textbf{88.56}} \\

&\multicolumn{1}{|c}{ERAT$_2$}     &{41.54}   &{15.58} &{67.64} &{\underline{86.78}} &{\textbf{87.55}}  &{\underline{84.92}}   &{\underline{88.13}} \\

&\multicolumn{1}{|c}{ERAT w/ CE}  &{80.34}   &{65.39}  &{84.28} &{80.37} &{73.9}  &{78.67}   &{80.12} \\

&\multicolumn{1}{|c}{ERAT w/o RB}  &{84.44}   &{74.33} &{84.06} &{84.02} &{82.04}  &{78.46}   &{85.37} \\

&\multicolumn{1}{|c}{ERAT w/o Uni}  &{\underline{84.75}}   &{\underline{75.95}}  &\underline{84.67} &{85.41} &{\underline{87.46}}  &{\textbf{87.2}}   &{87.62} \\

&\multicolumn{1}{|c}{ERAT w/o Aug} &{{72.62}} &{{72.76}}  &{{71.19}} &{69.19} &{{70.17}} &{{69.99}} &{{70.61}}  \\

&\multicolumn{1}{|c}{ERAT w/o AT} &{{38.58}} &{{19.79}}  &{{53.71}} &{{66.15}} &{{12.54}} &{{11.86}} &{{81.31}} \\

&\multicolumn{1}{|c}{ERAT w/o SS} &{{48.66}} &{{8.49}}  &{{73.33}} &{{55.74}} &{{62.79}} &{{51.96}}  &{{55.69}}   \\\hline

\multirow{5}{*}{\tabincell{c}{\emph{Asymm}. \\\emph{$\kappa = 0.4$}}}
&\multicolumn{1}{|c}{ERAT}  &\bf{85.87}   &{75.31}  &\underline{87.99} &\bf{90.63} &{\underline{86.89}}  &{\textbf{85.08}}   &{90.46} \\

&\multicolumn{1}{|c}{ERAT$_{\infty}$}    &{83.18}   &{74.86} &{85.83} &{88.41} &{50.37}  &{30.08}   &{\underline{91.66}} \\

&\multicolumn{1}{|c}{ERAT$_2$}     &{51.2}   &{65.07} &{74.93} &{\underline{90.43}} &{83.74}  &{83.0}   &{\textbf{91.73}} \\

&\multicolumn{1}{|c}{ERAT w/ CE}  &{79.71}   &{65.44} &{82.55} &{82.69} &{78.34}  &{63.58}   &{84.3} \\

&\multicolumn{1}{|c}{ERAT w/o RB}  &{\underline{85.53}}   &\underline{78.67} &{87.8} &{85.83} &{80.11}  &{77.93}   &{78.28} \\

&\multicolumn{1}{|c}{ERAT w/o Uni}  &{{85.28}}   &{\textbf{81.69}} &\bf{89.23} &{89.69} &{\bf{88.87}}  &{\underline{83.09}}   &{90.22} \\

&\multicolumn{1}{|c}{ERAT w/o Aug} &{{78.88}} &{{74.01}} &{{74.61}}  &{{72.93}} &{{77.7}} &{{77.52}}  &{{76.81}}\\

&\multicolumn{1}{|c}{ERAT w/o AT} &{{45.53}} &{{23.13}}  &{{57.98}} &{{71.16}} &{{10.39}} &{{12.87}} &{{83.24}} \\

&\multicolumn{1}{|c}{ERAT w/o SS} &{{67.22}} &{{72.66}}  &{{65.73}} &{{70.49}} &{{66.48}} &{{66.06}} &{{73.51}}   \\\hline

\toprule[1pt]
\end{tabular}
}
\end{table}

Here, we test the importance of key components of the ERAT. To this end, we introduce four variants. The first two are single-defense methods, i.e., ERAT$_{\infty}$ and ERAT$_{2}$, which train the model with one type of imaginary data perturbations, i.e., $\ell_{\infty}$-norm and $\ell_{2}$-norm bounded perturbations. The third (i.e., ``ERAT w/ CE'') is to use the conventional CE loss as the scoring function, which only measures how much the model agrees with the ground-truth. The fourth is not to consider class rebalancing (i.e., ``ERAT w/o RB''). The fifth is the variant that does not use uniform sampling for training, i.e., ``ERAT w/o Uni''. The sixth, i.e., ``ERAT w/o Aug'', is the variant that does not use strong augmentation for training. The variant ``ERAT w/o AT'' is the one that does not adopt adversarial training. The last is the variant that does not use sample selection for training, i.e., ``ERAT w/o SS''. From the results on CIFAR-10 recorded in Table \ref{alba}, we can observe that adversarial training makes great contributions to robustness against data poisoning. Single-defense methods cannot perform well without knowing the perturbation type in the dataset. When considering different types of perturbations during the training, the proposed method can achieve comprehensive protection. The proposed sample selection including the scoring and class rebalancing strategy contributes significantly to the performance of the proposed method. Uniform sampling would not significantly influence the final performance compared to the one that trains the model with all perturbations for each data. Strong augmentation can improve the generalization ability of the proposed method. All these demonstrate the effectiveness of each component of the proposed ERAT.

We also analyze the influence of the rebalancing degree by varying the value of $\eta$, where the results on CIFAR-10 and CIFAR-100 are reported in Figure \ref{eta_acc}. With $\eta$ decreasing, the balancing degree would increase. When $\eta =0$, an equal number of data would be selected to construct the labeled set, which is equal to the uniform sample selection proposed by \cite{karim2022unicon}.  When $\eta =1$, the selection would not be rebalanced and all small-loss samples would be deemed as clean. From the results, it can be observed that when $\eta=0.1$, the proposed method can achieve the best balance between the quality of clean label selection and the degree of class balancing.

The effect of $\lambda$ is tested, where the results of $\lambda = 0, 5, 10, 15, 25, 30, 50 \cdot clip(t_e/16, 0,1)$ are reported in Figure \ref{lam_acc}. From the results, we can observe that without pseudo-labeling to guide the learning of unlabelled data, the performance of the model would degrade. Meanwhile, too-small or too-large weights of training with unlabelled data would negatively influence the effectiveness of the model.

\subsection{\textbf{RQ5}: Efficiency}

The proposed method adopts the uniform sampling operation to reduce the computational costs of training with adversarial examples. For example, on CIFAR-10, the training time for the proposed method is about 17.8 hours, which is far below the time of training with all perturbations for each data, i.e., about 28.1 hours. From the results in Table \ref{alba}, the proposed uniform sampling operation does not significantly hurt the performance. This demonstrates the efficiency and effectiveness of the method. 

\section{Conclusions and future work}

In this paper, we investigate the significant threats of data and label corruptions to model training, and propose an Effective and Robust Adversarial Training (ERAT) method. To defend against data perturbations, the ERAT performs adversarial training between a noise generation module and the predictive model, where the noise generation module produces vicious poisoned data to improve model resilience and generalization ability. To avoid computation ahead and accelerate training, we uniformly sample attacking models from the perturbation set to generate noisy data. At the same time, a scoring-based class-rebalancing sample selection strategy is designed to effectively differentiate clean labels from noisy labels. Semi-supervised learning is accordingly performed to train the model to maintain the semantic consistency between data of different augmentation perturbation views. As a result, the negative effects of dual corruption can be significantly alleviated. Through extensive experiments, we verify the superiority of the proposed method.

However, this work also has some limitations and needs further improvement. First, adversarial training can only empirically demonstrate the effectiveness of defending against data corruption. How to theoretically ensure the robustness still remains a challenge. In addition, adversarial example generation requires additional computational costs. Improving efficiency is an important research direction. Second, following previous data corruption work, this work only considers two types of data perturbations, i.e., $l_2$ and $l_\infty$. However, there are more types in the literature, e.g., $l_0$ and $l_1$-norm, \textbf{SSIM} \cite{gragnaniello2021perceptual},  \textbf{LPIPS} \cite{zhang2018unreasonable}, and Wasserstein based data perturbations \cite{wu2020stronger}, despite they were not originally developed for data corruption. Studying their influence on the robustness of models and accordingly designing resilient methods deserve more attention. Lastly, this work is conducted based on the assumption that data and label corruption processes are independent of each other. Nevertheless, there might be a scenario in which data and label corruptions are deployed simultaneously. Exploiting this case would help us further test the robustness of existing methods and encourage more effective defense methods to ensure robustness and practicability.  

{\small
\bibliographystyle{IEEEtran}
\balance
\bibliography{IEEE}
}

\end{document}